\documentclass[conference]{IEEEtran}
\IEEEoverridecommandlockouts
\usepackage{cite}
\usepackage{adjustbox}
\usepackage{amssymb,lineno}
\usepackage{algorithm}
\usepackage{algorithmic}
\usepackage{multirow}
\usepackage[tight,footnotesize]{subfigure}
\usepackage{epsfig}
\usepackage{amssymb}
\usepackage{footmisc}
\usepackage{footnpag}
\usepackage{amsmath}
\usepackage{paralist}
\usepackage{color}
\usepackage[dvipsnames]{xcolor}
\usepackage{perpage}
\usepackage{verbatim}
\usepackage{enumitem}
\usepackage{tikz}
\usepackage{listings}
\usepackage{graphicx}

\def\BibTeX{{\rm B\kern-.05em{\sc i\kern-.025em b}\kern-.08em
    T\kern-.1667em\lower.7ex\hbox{E}\kern-.125emX}}

\setlist[itemize]{topsep=\parskip}

\begin{document}

\newcommand\xin[1]{{\color{cyan}{Xin: #1}}}
\newcommand\sheng[2]{{\color{red}{Sheng: #1}}}

\pagestyle{plain}

%\title{An Autoencoder based Error-bounded Lossy Compressor for Scientific Datasets}
\title{Exploring Autoencoder-based Error-bounded Compression for Scientific Data}

\author{\IEEEauthorblockN{Jinyang Liu,\IEEEauthorrefmark{1}
Sheng Di,\IEEEauthorrefmark{2}
Kai Zhao,\IEEEauthorrefmark{1}
Sian Jin,\IEEEauthorrefmark{3}
Dingwen Tao,\IEEEauthorrefmark{3} 
Xin Liang,\IEEEauthorrefmark{4}
Zizhong Chen,\IEEEauthorrefmark{1}
Franck Cappello\IEEEauthorrefmark{2}\IEEEauthorrefmark{5}}
\IEEEauthorblockA{\IEEEauthorrefmark{1}University of California, Riverside, CA, USA}
\IEEEauthorblockA{\IEEEauthorrefmark{2}Argonne National Laboratory, Lemont, IL, USA}
\IEEEauthorblockA{\IEEEauthorrefmark{3}Washington State University, Pullman, WA, USA}
\IEEEauthorblockA{\IEEEauthorrefmark{4}
Missouri University of Science and Technology, Rolla, MO, USA}
\IEEEauthorblockA{\IEEEauthorrefmark{5}
University of Illinois at Urbana-Champaign, Urbana, IL, USA}
jliu447@ucr.edu, sdi1@anl.gov, kzhao016@ucr.edu, sian.jin@wsu.edu, \\dingwen.tao@wsu.edu, xliang@mst.edu, zizhong.chen@ucr.edu, cappello@mcs.anl.gov
\thanks{Corresponding author: Sheng Di, Mathematics and Computer Science Division, Argonne National Laboratory, 9700 Cass Avenue, Lemont, IL 60439, USA}
}

\maketitle

\begin{abstract}
Error-bounded lossy compression is becoming an indispensable technique for the success of today's scientific projects with vast volumes of data produced during simulations or instrument data acquisitions. Not only can it significantly reduce data size, but it also can control the compression errors based on user-specified error bounds. Autoencoder (AE) models have been widely used in image compression, but few AE-based compression approaches support error-bounding features, which are highly required by scientific applications. To address this issue, we explore using convolutional autoencoders to improve error-bounded lossy compression for scientific data, with the following three key contributions. (1) We provide an in-depth investigation of the characteristics of various autoencoder models and develop an error-bounded autoencoder-based framework in terms of the SZ model. (2) We optimize the compression quality for the main stages in our designed AE-based error-bounded compression framework, fine-tuning the block sizes and latent sizes and also optimizing the compression efficiency of latent vectors. (3) We evaluate our proposed solution using five real-world scientific datasets and compare them with six other related works. Experiments show that our solution exhibits a very competitive compression quality among all the compressors in our tests. In absolute terms, it can obtain a much better compression quality (100\%$\sim$800\% improvement in compression ratio with the same data distortion) compared with SZ2.1 and ZFP in cases with a high compression ratio.        
\end{abstract}

%\begin{IEEEkeywords}
%Error-bounded lossy compression, autoencoder, scientific datasets
%\end{IEEEkeywords}

\section{Introduction}
\label{sec:introduction}

Today's scientific applications are producing extremely large amounts of data during simulation or instrument data acquisition. Advanced instruments such as the Linac Coherent Light Source (LCLS) \cite{lcls-ii} and Advanced Photon Source  \cite{APSU}, for example, may produce vast amounts of data with a very high data acquisition rate (250 GB/s \cite{use-case}).  Consequently, reducing the data volumes with user-tolerable data distortion is critical to efficient data storage and transfer. %Community Earth System Model (CESM)\cite{cesm}, for instance, can generate nearly 2.5 PB of data \cite{glecler} for the Coupled Model Intercomparison Project (CMIP) 5, which may produce 170 TB of analysis data submitted to the Earth System Grid \cite{esg}. 

%and these data need to be transferred from the instruments to remote supercomputers through a relatively slow network (1GB/s in general \cite{} and up to $\sim$ 25GB/s \cite{use-case} even in a dedicated network). As such, significantly reducing the data volume is significant to today's scientific projects.  

Error-bounded lossy compression is arguably the most efficient way to significantly reduce the data volumes for scientific applications with big data issues. Unlike lossless compressors \cite{gzip,zstd,zlib,fpc} that suffer from very low compression ratios (generally $\sim$2:1) on floating-point datasets, error-bounded lossy compressors can obtain fairly high compression ratios (10+ or even several hundred \cite{sz16,sz17,use-case}). Moreover, error-bounded lossy compressors are able to keep a high fidelity of the reconstructed data for the user's post hoc analysis based on the user's required bounds on data distortion, as verified by many recent studies \cite{cesm-eval-hpdf14,ssem,Baker-Climate17}. 

%General-purpose state-of-the-art error-bounded lossy compressors (such as ZFP \cite{zfp}, FPZIP \cite{lindstrom2006fast}, and SZ \cite{sz16,sz17}) often suffer from significant data distortions when the user requires a fairly high compression ratio.
Error-bounded lossy compressors can be split into two models: prediction-based and transform-based models. Prediction-based compressors (such as SZ \cite{sz16,sz17}) may suffer from low reconstructed data quality at high compression ratios because they have to predict each data point using the reconstructed data values nearby instead of original data, in order to guarantee the bounded errors during the decompression. To obtain a high compression ratio, therefore, one has to set the error bound relatively large; and as a result, the data prediction accuracy can be degraded significantly because of large errors in the reconstructed data, leading to limited compression ratios in turn. For transform-based compressors (such as ZFP \cite{zfp}), a large compression ratio means that a very limited number of coefficients or bit-planes can be preserved for reconstructing the data and thus will considerably lower the data reconstruction quality.   

As a classic type of deep learning model, the Autoencoder (AE) has been gaining more and more attention. Such a deep neural network (DNN) architecture is composed of both an encoder (encoding the input data) and a decoder (decoding the encoded data) and is trained to minimize the error between the reconstructed data and the initial data. In general, because the trained encoder and decoder can be used separately, AE can be used to learn efficient data representation (or coding), typically for  dimensionality reduction. The corresponding DNN will be trained to reconstruct the main patterns in the dataset effectively based on the reduced information generated from the original data. Recently, several variations of the AE have been developed with different model frameworks and training paradigms for improving the effectiveness of data reconstruction and for handling more tasks such as data generation.
Nevertheless, although AE has been widely used in the image compression domain, few studies explored the possibility of leveraging it for error-bounded compression models for scientific datasets.   

In this paper, we explore the possibility of leveraging the AE model to improve the error-bounded lossy compression significantly. Such a study faces several challenges. First,  many types of autoencoders exist, each with different architectures or training methods, so that determining the most effective AE model is challenging. Second, applying  AE in the error-bounded model with a proper configuration setting is nontrivial. Third, latent vectors from AE need to be stored in the compressed data, so minimizing the latent vector overhead while maintaining a high reconstruction quality is challenging and critical to getting a good rate distortion in high compression ratio cases. 

In this work, we propose a novel error-bounded lossy compressor,  AE-SZ, which combines the classic prediction-based error-bounded compression framework SZ \cite{sz16,sz17} and the Sliced-Wasserstein Autoencoder (SWAE) model with convolutional neural network implementations.
The key contributions of the paper are summarized as follows:
\begin{itemize}
\item Our autoencoder-based error-bounded compression framework is designed on the basis of a blockwise model, which can adapt to diverse data changes in a dataset well.
To the best of our knowledge, AE-SZ is the first AE-based error-bounded lossy compressor that exhibits a better rate distortion than the three state-of-the-art models SZauto \cite{Kai-HPDC2020}, SZ \cite{Xin-bigdata18}, and ZFP \cite{zfp}. 
\item We investigate various autoencoder models and identify the most effective one for the error-bounded lossy compression model and also carefully optimize the related configurations, such as block sizes and strategies of compressing latent vectors. 
\item We evaluate the proposed AE-SZ by using the scientific datasets generated by five different real-world high-performance computing (HPC) applications across different domains. We identify the effectiveness of AE-SZ by comparing it with two other AE-based lossy compression methods and four other state-of-the-art error-bounded lossy compressors. Our experiments show that  AE-SZ is the best compression method in the category of AE-based compressors. AE-SZ also exhibits competitive rate distortion compared with existing state-of-the-art error-bounded lossy compressors. Specifically, when the compression ratio is greater than 100, AE-SZ can get 100\%$\sim$800\% higher compression ratios than SZ2.1 and ZFP, with the same peak signal-to-noise ratio (PSNR).   
\end{itemize}

The rest of this paper is organized as follows. In Section~\ref{sec:related} we discuss related work. In Section~\ref{sec:problem} we formulate the research problem. %In Section \ref{sec:analysis}, we perform an in-depth characterization for many DNN models that can be used in the data reduction. 
In Section~\ref{sec:designoverview} we present the overall design of AE-SZ as well as the detailed optimization strategies. In Section~\ref{sec:evaluation} we evaluate our solution using multiple real-world scientific simulation datasets. In Section \ref{sec:conclusion} we conclude the paper with a vision of the future work.  
\section{Related Work}
\label{sec:related}

Error-bounded lossy compression techniques have been studied for years since lossless compression suffers from very low compression ratios (generally 2:1 \cite{son2014data}). Many error-bounded lossy compressors have been developed for compressing scientific datasets \cite{zfp, lindstrom2006fast, wavesz,ainsworth2019multilevel, li2017achieving, delaunay2019evaluation, zender2016bit, lakshminarasimhan2013isabela, quantum, Pastri,sz-psnr,cusz,8257962,DeepSZ,liang2018efficient}. These can be categorized into prediction-based models (e.g., SZ \cite{sz16,sz17} and FPZIP  \cite{lindstrom2006fast}) and transform-based models (e.g., ZFP \cite{zfp}). Among these compressors, SZ2.1 \cite{sz16,sz17,Xin-bigdata18} and ZFP \cite{zfp} are the two main state-of-the-art works with wide public usage. Several works have also been developed based on SZ2.1. For example, SZauto \cite{Kai-HPDC2020} merges second-order regression/Lorenzo and automatic parameter tuning in the SZ framework, and SZinterp \cite{sz_interp} applies dynamic spline interpolation into data prediction and achieves significant improvement in the prediction accuracy. For assessing the lossy compressors, \cite{z-checker} provides an effective framework.
%FPZIP \cite{}, for example, is a typical prediction-based lossy compressor including four steps: predicting each point's value by Lorenzo, mapping original values and predicted values to sign-magnitude binary integer representations, then computing the delta of the representations, followed by an entropy encoding. 

%SZ \cite{sz16,sz17}, is a typical prediction-based lossy compressor, which involves four key steps: (1) predicting data by a hybrid predictor \cite{liang2019significantly}, (2) performing a linear-scale quantization \cite{sz17} to control errors, (3) conducting a customized Huffman encoder, and (4) performing a dictionary encoder (Zstd \cite{zstd} by default) to significantly reduce the data size. ZFP \cite{zfp} is a typical transform-based compressor, which applies an optimized orthogonal block-wise transform on the exponent-aligned data, followed by a customized embedded encoding algorithm. Based on our prior studies \cite{fraz, sz16, sz17, Xin-bigdata18}, SZ and ZFP exhibits the best rate-distortion in most of cases from among all the existing state-of-the-art lossy compressors, so we mainly compare our AE-based lossy compressor to these two compressors in our experiments. 

With the fast growth of deep learning, a recent research trend is leveraging deep learning models such as autoencoders on data compression tasks. Successful works for AE-based image compression include \cite{theis2017lossy,cheng2018deep,zhou2018variational,chen2019neural,balle2017endtoend,balle2018variational,johnston2019computationally}, which design different convolutional networks for image feature extracting and reconstruction and combine them with quantization and encoding algorithms. Unlike scientific lossy compressors, however, those autoencoder-based image compression models are not designed to compress floating-point data and do not provide a strict error-controlling scheme based on scientific user’s requirements on post hoc analysis.
Recently,  a few works have used an autoencoder to compress scientific data. Glaws et al. \cite{glaws2020deep} presented a convolutional autoencoder for lossy compression of turbulence flow simulation data (with a fixed compression ratio of 64). The authors proposed an AE model including 12 residual blocks (i.e., skip connections \cite{he2016deep}) to extract features and 3 compression layers to reduce features in both the encoder and decoder. However, different from our work which can provide a strict control of local error (e.g., relative/absolute error) and can be adapted to any scientific datasets, this AE-based scientific compressor is not error-bounded and is designed only for turbulence data. The fixed compression ratio is also a limitation.

Choi et al. \cite{choineural} proposed another specific variational autoencoder approach for physics plasma simulation data compression. The proposed AE model focuses on minimizing loss of information under physics constraints (e.g., mass, energy, moment) by adopting physics-informed optimization functions and refinement layers. However, unlike this work using quantized latent vectors (integer-based), our solution applies lossy compression to floating-point latent vectors, which provides a highly flexible tradeoff between compression and accuracy. In \cite{choineural} the authors presented a brief version that lacks some important details for replication of their experiments, so we did not compare their performances in our paper.

Liu et al. \cite{liu2021high} developed an autoencoder method for scientific data compression. Their proposed AE model includes three fully connected layers for both encoder and decoder (i.e., a total of seven layers including the latent vector), and the size of each layer is reduced/increased by $8\times$ compared with its previous layer, thus leading to an overall compression ratio of $512\times$. The limitations of this work are that the autoencoder in their model processes only 1-D data, and the experiments are also based mainly on small-scale 1D scientific data. Our AE-SZ framework overcomes the limitations by designing and training convolutional autoencoders that are aware of dimensional information and are well adapted to large-scale data with relatively high prediction speed and accuracy.
\section{Background and Problem Formulation}
\label{sec:problem}

In this section, we describe the research background and formulate the research problem.

\subsection{Research Background -- Autoencoder}

%We use Figure \ref{fig:aebasic} to illustrate a stereotype autoencoder model. Taking the input, an encoder network encodes it to a latent vector in reduced size, and the decoder network decodes the latent vector to an approximate reconstruction of the original input. The latent vector (named as Z in the figure), standing as a compressed representation of the input data, can be 1-D vectors or multi-dimensional tensors, and different autoencoder types have different technical details for computation of the latent vector. %For example, variational autoencoder (VAE) \cite{kingma2013auto} first computes means and variances of latent vectors, and then samples the latent vector randomly from a prior distribution.  

%\begin{figure}[ht]
%  \centering
%  \raisebox{-1mm}{\includegraphics[scale=0.64]{figures/aebasic.png}}
%  \vspace{-1mm}
%  \caption{Design Overview}
%  \label{fig:aebasic}
%\end{figure}

We describe autoencoder briefly as follows. A stereotype autoencoder model is composed of an encoder network and a decoder network. The former encodes the input data to a latent vector in reduced size, and the latter decodes the latent vector to an approximate reconstruction of data. The latent vector stands as a compressed representation of the input data, and different autoencoders have different technical details for the computation of the latent vector.

\begin{figure}[ht]
  \centering
  \raisebox{-1mm}{\includegraphics[scale=0.8]{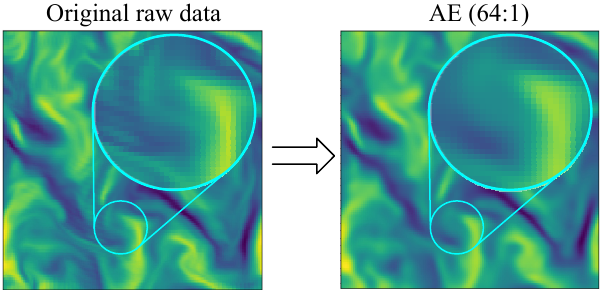}}
  \vspace{-1mm}
  \caption{Reconstructed data of AE  (64$\times$) on a turbulence dataset \\(original value range: [$-$3.06 , 2.64], max pointwise absolute error = 1.2)}
  \label{fig:ae64-vis}
\end{figure}

The nature of the autoencoders grants them the potential for being leveraged for data reduction, in that the reconstructed data based on the latent vector can approximate the original data to a certain extent. Figure \ref{fig:ae64-vis} shows the visualization of the reconstructed data versus the original data with the autoencoder \cite{glaws2020deep} (reduction ratio = 64$\times$) on a turbulence dataset.

As a historical and well-researched neural network model,  multiple variations have been proposed for the AE model. 
%The key difference among the variations is the design of loss function, also known as the optimization target. In this term, most recent works focus on mapping the latent space to predefined probability distribution spaces, which allows the decoder network part to work separately as a data generator. This feature may also improve the data compression and reconstruction capability, as it makes the model learn more information of the probability distribution of the original data and also alleviates overfitting. 
In what follows, we mainly present SWAE, which is to be used as the fundamental AE model in our designed AE-SZ compressor. 

SWAE \cite{kolouri2018sliced} is a derivation of Wasserstein Autoencoder (WAE) \cite{tolstikhin2017wasserstein}, and regularizes the autoencoder loss with the sliced-Wasserstein distance between the distribution of the encoded training samples and a predefined samplable distribution. From \cite{kolouri2018sliced}, marking $\phi$ as the encoder and $\psi$ as the decoder, given a latent dimension $d$, a regularization coefficient $\lambda$, a number of random projections $L$, and a predefined latent distribution $q_Z$, SWAE optimizes the following loss function:
\begin{equation}
\label{eq:swaeloss}
\begin{array}{l}
\hspace{-3mm}\mathcal{L}(\phi, \psi)=\frac{1}{M} \sum_{m=1}^{M} c\left(x_{m}, \psi\left(\phi\left(x_{m}\right)\right)\right)\\
\hspace{11mm}+\frac{\lambda}{L M} \sum_{l=1}^{L} \sum_{m=1}^{M} c\left(\theta_{l} \cdot \tilde{z}_{i[m]}, \theta_{l} \cdot \phi\left(x_{j[m]}\right)\right) \hspace{-3mm},
\end{array}
\end{equation}

in which

$\left\{x_{1}, \ldots, x_{M}\right\}$ is sampled from training set (i.e. $p_X$),

$\left\{\tilde{z}_{1}, \ldots, \tilde{z}_{M}\right\}$ is sampled from $q_Z$,

$\left\{\theta_{1}, \ldots, \theta_{L}\right\}$ is sampled from $\mathrm{S}^{d-1}$ (K-dimensional unit sphere),

$i[m]$ and $j[m]$ are the indices of sorted $\theta_{l}\cdot\tilde{z}_{m}\mathrm{~s}$ and $\theta_{l}\cdot\phi\left(x_{m}\right)$, respectively, and 
\begin{equation}
c(x,y)=||x-y||_2^2  .
\end{equation}

Kolouri et al. \cite{kolouri2018sliced} proved that optimizing this loss function is equal to optimizing

\begin{equation}
\operatorname{argmin}_{\phi, \psi} W_{c}\left(p_{X}, p_{Y}\right)+\lambda S W_{c}\left(p_{Z}, q_{Z}\right) ,
\end{equation}
in which $W_{c}\left(p_{X}, p_{Y}\right)$ is the Wasserstein distance from $p_{X}$ (distribution of input data $X$) to $p_{Y}$ (distribution of decoded data $Y$) and $S W_{c}\left(p_{Z}, q_{Z}\right)$ is the sliced-Wasserstein distance from $p_{Z}$ (distribution of encoded latent $Z$) to $q_{Z}$. Kolouri et al. \cite{kolouri2018sliced} also show the efficiency of computing Eq. \ref{eq:swaeloss}.

\subsection{Problem Formulation}
\subsubsection{Leveraging Autoencoders in Error-bounded Scientific Data Compression}
The autoencoder itself cannot bound the compression errors, which is a significant gap to scientific user's demand for error controls. As shown in Figure \ref{fig:ae64-vis}, the maximum point-wise compression error is up to 1.2, which is about 20\% of the original data value range ($-$3.06, 2.64]. In comparison, scientists often need to control the point-wise errors to a much smaller bound such as 1\% of the original value range \cite{sz17,Xin-bigdata18}. 

In this work we aim to develop a deep-learning based error bounded lossy compressor. 
%Given a scientific dataset (denoted by $D$=\{$d_i$, $i$=1,2,$\cdots$,$N$\}) which includes $N$ floating-point data values (a.k.a., data points), and a user-specified absolute error bound (denoted by $e$), the objective is to develop a deep learning based lossy compressor that can obtain a high compression ratio with a user-acceptable data distortion level.
%higher compression ratio than the state-of-the-art error-bounded lossy compressors (such as SZ and ZFP) do, with the same data distortion level. 
Specifically, for some scientific applications, we train neural networks based on a certain amount of training data, and then apply the trained networks to compress the testing data generated by the same applications. We separate the training data and testing data because we expect that the pre-trained networks can be used to compress new data for the same applications, such that the training time and model size can be excluded from the compression time and size. In our experiments, the training and test data are from different time steps or the simulation running with different configuration settings in the same application.   
%The compressed data includes the core tensor (e.g., latent vector of the autoencoder) and other encoded bytes to control the error bound (such as quantization codes).
\subsubsection{Math Formulations for Error-bounded Lossy Data Compression}
The compression ratio (denoted by $\rho$) is defined as $\frac{|D|}{|D'|}$, where $|D|$ and $|D'|$ denote the original data size and compressed data size (both in bytes), respectively.  

Error-bounded lossy compression has one important constraint, namely, that the reconstructed data respect a user-specified error bound (denoted by $e$) strictly. Under this constraint, the rate-distortion often serves as a criterion to assess the compression quality, which involves two critical terms: bit rate and data distortion. The bit rate is defined as the average number of bits used to represent one data point after the compression; hence, the lower the bit rate, the higher the compression ratio. In the lossy compression community rate distortion is often evaluated by the PSNR, defined as shown below: 
\def\formulaPSNR{
P\hspace{-0.3mm}S\hspace{-0.3mm}N\hspace{-0.3mm}R = 20\log_{10}{vrange(D)} \hspace{-0.3mm}-\hspace{-0.3mm} 10\hspace{-0.3mm}\log_{10}{mse(D,\hspace{-0.3mm}D')} \hspace{-0.1mm}
}
\vspace{-2mm}
\begin{equation}
\label{eq:psnr}
  \formulaPSNR
  \vspace{-1mm},
\end{equation}
\noindent
where $D'$ is the reconstructed dataset after decompression (i.e., decompressed dataset), vrange($D$) represents the value range of the original dataset $D$ (i.e., the difference between its highest value and lowest value), and \emph{mse} refers to mean squared error. 
The higher the PSNR value is, the smaller the mean squared error, which means higher precision of the decompressed data. 

Our objective is to obtain higher compression ratios than other related works (including other deep-learning-based compressors and traditional error-bounded lossy compressors) with the same PSNR value, while also strictly respecting the user's error bound, especially aiming at optimizing the use cases with high compression ratio. We can write the research problem formulations as follows:

\def\formulaEBFORM{
\begin{array}{l}
 maximize\hspace{2mm}\rho  \\ 
 s.t. \hspace{4mm}PSNR(D,D') = \lambda \\
 \hspace{8mm}\hspace{1mm}|d_i  - d_i'| \leq e
 \end{array}
}
\vspace{-2mm}
\begin{equation}
\label{eq:eb-form}
  \formulaEBFORM  ,
\end{equation}
where $\lambda$ is a particular PSNR value representing a specific data distortion level and $d_i$ and $d_i'$ refer to any data point in the original dataset $D$ and decompressed dataset $D'$, respectively.

\section{AE-SZ: Autoencoder-based Error-bounded Lossy Compression Framework}
\label{sec:designoverview}

In this section, we present the design overview of AE-SZ and describe the detailed optimization strategies for AE-SZ. 

\subsection{Design Overview of AE-SZ}

\begin{figure}[ht]
  \centering
  \raisebox{-1mm}{\includegraphics[scale=0.58]{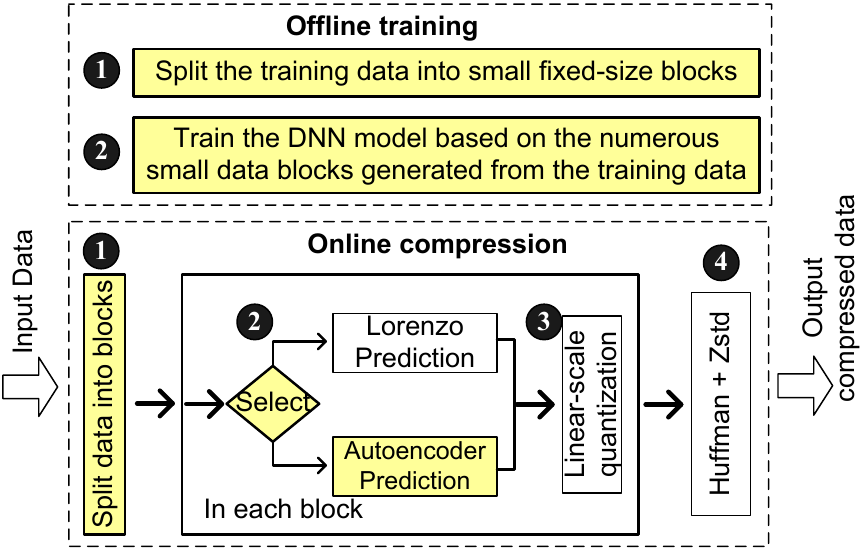}}
  \vspace{-1mm}
  \caption{Design overview of AE-SZ (highlighted parts include our specifically optimized design compared with SZ compressor)}
  \label{fig:overview}
\end{figure}

We present the overall framework of our designed autoencoder-based error-bounded lossy compression framework AE-SZ as shown in Figure \ref{fig:overview}. The overall compression involves two stages: offline training and online compression.

During the offline training, we split the training data snapshots into multiple small fixed-size blocks (such as 32$\times$32 for a 2D data field or 8$\times$8$\times$8 for a 3D data field) and train the network with numerous small blocks. The advantage of such a design is twofold: (1) the AE model works more efficiently on the divided data blocks, which can catch fine-grained data features; (2) such a data-splitting design creates numerous training samples (i.e., data blocks), so that the AE model is tractable.    

During the online compression, AE-SZ executes four steps as shown in Figure \ref{fig:overview}: (1) splitting the input data to be compressed into many small blocks (with the same block size as during the training stage), (2) prediction, (3) linear-scale quantization,  and (4) entropy/dictionary encoding. Specifically, in each block, the data are predicted by a predictor (either autoencoder or Lorenzo), and the prediction errors will be quantized based on the user's error bound, followed by Huffman encoding and Zstd \cite{zstd}. The Lorenzo predictor is similar to the one used in SZ2.1. Specifically, under the Lorenzo predictor, the $i$th data point $d_i$ is predicted by three nearby data values in 2D data ($d_{i,j}$$\leftarrow$$d_{i,j-1}$ + $d_{i-1,j}$ $-$ $d_{i-1,j-1}$) 
or by 7 nearby values in 3D data ($d_{i,j,k}$$\leftarrow$$d_{i-1,j,k}$ + $d_{i,j-1,k}$ + $d_{i,j,k-1}$ $-$ $d_{i-1,j-1,k}$ $-$ $d_{i-1,j,k-1}$ $-$ $d_{i,j-1,k-1}$ + $d_{i-1,j-1,k-1}$). 
We refer the readers to read our prior work \cite{sz17} for more details. We note that the only difference between the Lorenzo predictor in AE-SZ and \cite{sz17} is that AE-SZ makes the selection between classic Lorenzo and mean-Lorenzo separately on each block instead of a global switching mechanism as in \cite{sz17}. That is, if a data block can be better predicted by its mean value than by classic Lorenzo, AE-SZ will use the mean value for prediction, and all the involved mean values will be saved losslessly. We find that this mean-Lorenzo predictor can make up for the deficiencies of classic Lorenzo and AE under extremely high error bounds (such as $\sim$1E-1). 

The compressed data generated by AE-SZ consists of three parts: a header containing metadata (with trivial space cost), lossy compressed latent vectors from autoencoders, and quantization bins (losslessly encoded). 

The main difference between AE-SZ and SZ2.1\cite{Xin-bigdata18} is that SZ2.1 includes two data predictors for compression: linear regression \cite{Xin-bigdata18} and Lorenzo \cite{lorenzo}, whereas AE-SZ replaces the linear regression predictor by a pre-trained autoencoder. 
%The reason for the replacement is autoencoders are advanced to linear regression for the capability of learning and reconstructing the non-linearity of input data. 
For scientific datasets in which the data changes could be diverse, autoencoders can overcome the limitation of linear regression, which can only approximate the data using flat hyperplanes. 

The pseudo-code of the AE-SZ compression procedure is presented in Algorithm \ref{alg:aesz}. As mentioned before, AE-SZ compresses the input data block by block, and the compression of each block follows the same routine. Thus, in the following, we describe the compression procedure mainly on a single data block (i.e., line 2$\sim$16), without loss of generality. 
For any block, AE-SZ first generates predicted data based on two predictors (Lorenzo and autoencoder) for this block respectively (line 3$\sim$8). Then, the predictor with lower element-wise $l$1-loss is selected for this block (line 9$\sim$13). The reason is that the smaller the prediction errors are, the more uneven the distribution of quantization bins in general, and hence the higher compression ratio of quantization bins. 
%We note that the predicted values need to be identical during the compression and decompression procedure for guaranteeing the controlled errors. 
Then, AE-SZ uses linear-scale quantization to quantize the prediction errors based on the user-specified error bound $e$ (line 14). Similar to SZ2.1 \cite{sz17,Xin-bigdata18}, we need to set a maximum number of quantization bins (65,536 by default) for the linear quantization, in order to keep high performance. The total quantization range may not cover all predicted values as the prediction errors may be large. The corresponding data points, called \emph{unpredictable} data, will be saved separately (denoted as $U$ in line 14). For more details about linear-scale quantization, we refer readers to read our paper \cite{sz17}.

\begin{algorithm}
\footnotesize
\caption{AE-SZ Compression Algorithm}
\label{alg:aesz}
\renewcommand{\algorithmiccomment}[1]{/*#1*/}
\begin{flushleft}
\textbf{Input:} Input data $D$, block size $S$, error-bound $e$, latent error-bound $e_l$.\\ %dimension information  
\textbf{Output:} Compressed data $D'$=\{$\hat{z}$, $\hat{Z}$, $U$\}.
\end{flushleft}

\begin{algorithmic}[1]
\STATE Split $D$ into blocks of Size S(1D), S$\times$S(2D), or S$\times$S$\times$S(3D). 
\FOR{(each block $B$ in the data)}
\STATE  $z$ $\leftarrow$ $Eec(B)$. \COMMENT{Encode $B$ with the encoder network $Eec$.}  
\STATE  $z'$ $\leftarrow$ $f$($z$,$e_l$). \COMMENT{Get decompressed latent vector $z'$ based on $e_l$} 
\STATE $B'=Dec(z')$. \COMMENT{Get Decoded $B'$ using decoder network $Dec$}
\STATE $loss_1=||B-B'||_1$. \COMMENT{Compute $l$1 loss of $B'$ vs. $B$.}
\STATE $B''=Lorenzo(B)$. \COMMENT{Predict $B$ with Lorenzo.}
\STATE $loss_2=||B-B'||_1$. \COMMENT{Compute $l$1 loss of Lorenzo predictor.}
\IF{$loss_2\leq loss_1$}
\STATE $B_p=B''$. \COMMENT{Select Lorenzo-predicted values}
\ELSE 
\STATE $B_p=B'$. \COMMENT{Select Autoencoder-predicted values} 
\ENDIF
\STATE $Q,U=Quantize(B,B_p,e)$. \COMMENT{linear-scale quantization with $e$, to get quantization codes $Q$ and unpredictable data $U$.}
\ENDFOR
\STATE Compress all saved coefficients from AE and Lorenzo.
%\STATE $\hat{z}$ $\leftarrow$ Compress all the latent vectors $z$ \COMMENT{see section \ref{latent vector compression}}.
\STATE $H$ $\leftarrow$ Huffman\_Encode($Q$). \COMMENT{Huffman encoding}
\STATE $\hat{Z}$ $\leftarrow$ \emph{Zstd}($H$). \COMMENT{Zstd compression}
\end{algorithmic}
\end{algorithm}

In the following subsections, we present several critical optimization strategies for AE-SZ, which are developed in terms of fundamental takeaways we summarized from our in-depth analysis or comprehensive experimental evaluation. 
%In the following text, we describe the detailed design motivation (e.g., why do we need to combine Lorenzo and autoencoder) and optimization strategies (e.g., how to design the neural network and tune parameter optimization).
\subsection{Design Detail: AE network structure in AE-SZ}
\label{sec:aestruct}
\begin{figure}[ht]
  \centering
  \raisebox{-1mm}{\includegraphics[scale=0.6]{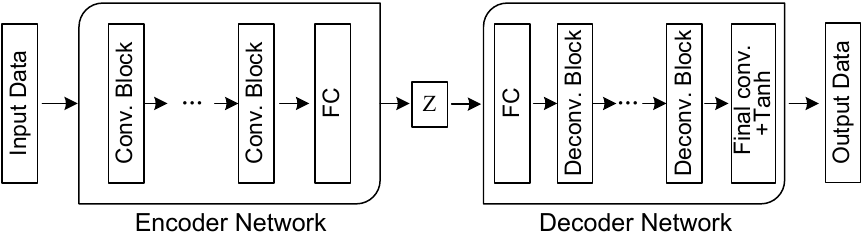}}
  \vspace{-4mm}
  \caption{Our Designed Blockwise Convolutional AE network for Compression}
  \label{fig:aeoverall}
\end{figure}

The structure of our designed autoencoder network used in AE-SZ is illustrated in Figure \ref{fig:aeoverall}. Like most autoencoders, it consists of an encoder network to generate the latent vectors as the compressed representation of input original data and a decoder network to reconstruct the data from latent vectors. The input of the network is (batches of) data blocks, which will be linearly normalized to the range of [-1, 1] based on the global maximum and minimum of data before being put in the network, and the output of the network needs to be denormalized to generate the final prediction values.

The encoder and decoder networks are both formed with several convolutional/deconvolutional blocks and a fully connected layer for resizing latents, and their structures are mirror-symmetric except for an additional final output layer-set in the decoder network.

\begin{figure}[ht]
  \centering
  \raisebox{-1mm}{\includegraphics[scale=0.8]{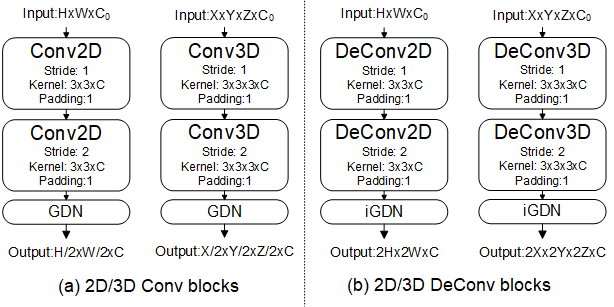}}
  \vspace{-1mm}
  \caption{(a) The Convolutional blocks used in AE-SZ encoder network. (b) The Deconvolutional blocks used in AE-SZ decoder network.}
  \label{fig:aeblocks}
\end{figure}

As shown in Figure \ref{fig:aeblocks}, the convolutional blocks in the encoder network are composed of the layer sequence of Convolution(Stride 1)-Convolution(Stride 2)-GDN, and the ones in the decoder network are composed of the layer sequence of Deconvolution(Stride 1)-Deconvolution(Stride 2)-iGDN. The size of each (de)convolutional kernel in the network is 3$\times$3 (2D case) or 3$\times$3$\times$3 (3D case). We take some experiments for the block design from the image compressive autoencoder in \cite{cheng2018deep} and \cite{zhou2018variational}. The reason we apply stride-1 convolutions before stride-2 convolutions is to increase the number of parameters of the network without fast reducing the size of feature maps. As reported in \cite{theis2017lossy} and \cite{cheng2018deep}, consecutively stacking stride-2 convolutions will harm the performance of the network.

In our AE-SZ autoencoder, we do not use traditional activation functions but use Generalized Divisive Normalization (GDN) \cite{balle2016density} as the activation function. In fact, according to Balle et al.'s work \cite{balle2018efficient}, GDN can provide better image reconstruction quality with trivial additional parameters compared with traditional activation and normalization functions such as Relu, LeakyRelu \cite{maas2013rectifier} and Batch Normalization \cite{ioffe2015batch}. Several existing lossy image compression autoencoder models \cite{balle2017endtoend,balle2018variational,johnston2019computationally,zhou2018variational,jia2019layered,xiao2019image} have leveraged GDN and proved its advantages. Our primary experiments also confirm that GDN outperforms other tested activation functions on scientific data lossy compression tasks. More details about GDN can be found in \cite{balle2016density} and \cite{balle2018efficient}. Following the common configurations, we apply the original GDN in convolutional blocks and apply its reverse iGDN in deconvolutional blocks.

To adapt to different datasets, the number of Convolutional blocks and the number of channels in each block may vary, but the overall structure remains the same. For example, The main difference between autoencoders used for 2D/3D datasets is just the dimension (2D or 3D) of convolutional/deconvolutional operation in the network layers (see Figure \ref{fig:aeblocks}). 

%There are two main reasons for AE-SZ not leveraging advanced network structures such as containing Residual Connection or Dense Connection. First, in scientific data compression tasks, in addition to compression bitrates corresponding to prediction accuracy, the compression time (aka., efficiency) is also vital for the framework. Large networks with complex structures harms the training and inference efficiencies. Second, in our early experiments, complex encoder/decoder networks such as ResNet networks can only make minimal contributions to the compression performances compared with sequential networks. To balance performance and efficiency, AE-SZ chooses a simple and still effective choice of network design.

The network model in AE-SZ is saved separately against the compressed data because it can be reused by different time steps or other simulations with different parameter settings, which is verified in our experiments (see Section \ref{sec:evaluation}).  
%As AE-SZ is designed for intensively generated and transferred scientific datasets from scientific simulations, unlike some previous works, the autoencoder does not need to be trained on data to be compressed, but is trained on another set of data snapshots of the same data field. 
%The training of the model for each data field needs only once, and the autoencoder can be used and updated for other data snapshots in the same data field, even for other runs of simulations. Therefore, we do not need save the model for each compression file, but just uses it as it is a part of the compressor itself and do not need to compute the compression ratio with the model size. 
%The experiments in the next sections will confirm the availability and effectiveness of this paradigm.

\subsection{Design Details: Choosing the Autoencoder Type}
\label{sec:swae}

\noindent
\textbf{Takeaway 1}: \textit{Sliced-Wasserstein Autoencoder is particularly suitable for data prediction in scientific data compression compared with other AE models.} 

A key point in designing AE-SZ is that we need to select the most appropriate model for scientific data prediction from multiple variations of autoencoder models. In AE-SZ, we select sliced-Wasserstein autoencoder (SWAE) for the AE compressor and predictor. The advantages of SWAE in data compression are as follows:
\begin{itemize}
    
\item Compared with the other tested autoencoders, SWAE shows less reconstruction loss on scientific data.

\item Different from traditional variational autoencoders (VAEs), the encoding and decoding computation in SWAE are both determinant. VAEs such as \cite{kingma2013auto,higgins2016beta,kumar2018dipvae,zhao2018infovae,chen2018logcosh} actually compute means and variances with input data and sample latent vectors with the means and variances from the prior distribution. Therefore, in multiple runs with the same input, the latent vector as the output of the encoder in a VAE will differ, which makes the VAE unstable for data compression tasks. %Unlike traditional VAEs which sample latent vectors from probability distributions, the sampling process in SWAE only appears in training loss computation.

\item Compared with Wasserstein autoencoders (WAE), the computation of training loss in SWAE is more numerically efficient. Similar to SWAE, WAE computes Wasserstein distances for training losses, and its computation cost is higher than the computation of sliced-Wasserstein distances. With both $n$ samples from the training set and prior distribution, the computational cost of the Wasserstein distance is $O(n^2)$ whereas the computational cost of the sliced-Wasserstein distance is $O(n\log{n})$.
\end{itemize}
Table \ref{tab:aetype} presents the reconstruction quality (PSNR) on different types of autoencoders that we explored. We trained 8 types of autoencoders on a split of snapshots of the CESM-CLDHGH data field:  a vanilla autoencoder, vanilla variational autoencoder \cite{kingma2013auto}, $\beta$-VAE \cite{higgins2016beta}, DIP-VAE \cite{kumar2018dipvae}, Info-VAE \cite{zhao2018infovae}, LogCosh-VAE \cite{chen2018logcosh}, WAE \cite{tolstikhin2017wasserstein}, and SWAE \cite{kolouri2018sliced}. After training, the AEs were tested by using another split of data snapshots. From this table, we observe that SWAE has the best prediction accuracy with the highest PSNR, which motivates us to use it as our final predictor in AE-SZ. 

\begin{table}[ht]
\centering
\footnotesize
  \caption {Average prediction PSNR of different types of autoencoders on CESM-CLDHGH data field} 
  \vspace{-2mm}
  \label{tab:aetype} 
  %\begin{adjustbox}{width=\columnwidth}
  \begin{tabular}{|c|c||c|c|}
  \hline
    \textbf{AE type}& \textbf{PSNR}&\textbf{AE type} & \textbf{PSNR}\\
    \hline
    AE &42.2 & Info-VAE & 26.5 \\
    \hline
    VAE &36.2 & LogCosh-VAE & 39.0 \\
    \hline
    $\beta$-VAE & 40.1 & WAE & 42.4 \\
    \hline
    DIP-VAE  & 32.2 & SWAE & \textbf{43.9}\\
    \hline
\end{tabular}
%\end{adjustbox}
\end{table}

\subsection{Design Detail: Optimizing AE Configurations}
\label{sec:aeconfig}

\noindent
\textbf{Takeaway 2}: \textit{The performance of AE may differ a lot with different configurations under the same model structure. Optimizing the AE configurations, especially the input block size and the latent vector size, is critical to the final performance of AE-SZ.}

As presented in Section \ref{sec:aestruct}, the AE network in AE-SZ has a flexible structure, which can accept various configurations such as different input block sizes, latent vector sizes, block numbers, and channel numbers. From those, the input block size and latent vector size are two critical hyperparameters for the performance,  corresponding to the scale of learned data patterns and the representation compactness of latent vectors. We need to optimize the input block size to have the best scope of data, and we need to optimize the latent vector size to balance prediction accuracy and latent overhead. 

Table \ref{tab:aeblocksize} shows the average prediction PSNR and AE-SZ compression ratio (error bound = 1E-2) of different input block sizes under the same latent ratio (input block size divides latent vector size, 64 for CESM-CLDHGH and 32 for NYX-baryon\_density). We conclude that optimizing the input block size is of great importance because autoencoders can achieve apparently different performances under the same latent overhead but various input block sizes. In our work, we optimize the input block size of the autoencoder in AE-SZ separately for each field, and we find that 32$\times$32 input block fits most of the 2D data fields tested and that 8$\times$8$\times$8 input block fits most of the 3D data fields tested.

\begin{table}[ht]
\centering
\footnotesize
  \caption {Average prediction PSNR and AE-SZ compression ratio (1e-2 error bound) of different input block sizes} 
  \vspace{-2mm}
  \label{tab:aeblocksize} 
  %\begin{adjustbox}{width=\columnwidth}
  \begin{tabular}{|c|c|c|c|c|c|}
  \hline
    \multirow{2}{*}{\textbf{Blocksize}}&\multicolumn{2}{|c|}{\textbf{CESM-CLDHGH}}&\multirow{2}{*}{\textbf{Blocksize}}& \multicolumn{2}{c|}{\textbf{NYX-baryon\_density}}\\
    \cline{2-3}\cline{5-6}
    &\textbf{PSNR}&\textbf{CR(1e-2)}&&\textbf{PSNR}&\textbf{CR(1e-2)}\\
    
    \hline
    16$\times$16 &42.5&55.5&8$\times$8$\times$8&\textbf{46.6}&\textbf{71.1}\\
    \hline
        32$\times$32 &\textbf{43.9}&\textbf{60.9}&16$\times$16$\times$16&35.7&23\\
    \hline
    64$\times$64 & 41.7 &50.1&32$\times$32$\times$32&28.9&23.9\\
    \hline

\end{tabular}
%\end{adjustbox}
\end{table}

Table \ref{tab:latentsize} presents the final compression ratio of AE-SZ under the error bound of 1E-2 with AEs of different latent sizes on the  Hurricane-U data field. The input block size is 8$\times$8$\times$8, and the rest part of the network remains the same for different latent sizes. We can see that different latent sizes bring a 40\%+ difference in final compression ratios, which motivates us to choose an appropriate latent size in our design. In what follows, we discuss how we reduce the latent vector size while maintaining high prediction accuracy of AEs.
%which can provide effective predictions without too much overhead.

\begin{table}[ht]
\centering
\footnotesize
  \caption {Compression ratio of AE-SZ under the error bound of 0.01 with AEs of different latent sizes on the Hurricane-U data field} 
  \vspace{-2mm}
  \label{tab:latentsize} 
  %\begin{adjustbox}{width=\columnwidth}
  \begin{tabular}{|c|c|c|}
  \hline
    \textbf{Latent size}&\textbf{Latent ratio}&\ \textbf{CR (1E-2)}\\
    \hline
    4 &128&123.4\\
    \hline
    6 &85.3&137.4\\
    \hline
    8 & 64& \textbf{149.1}\\
    \hline
    12  &45.7&127.7\\
    \hline
    16 & 32& 106\\
    \hline

\end{tabular}
%\end{adjustbox}
\end{table}

\subsection{Design Detail: Lossy compression of AE latent vectors}
\label{sec:latentcompression}

\noindent
\textbf{Takeaway 3}: \textit{Predicting the data with error-bounded lossy decompressed latent vectors can maintain a very small loss of prediction accuracy, while greatly reducing the latent vector size (i.e., latent overhead).}

\begin{figure}[ht]
  \centering
  \raisebox{-1mm}{\includegraphics[scale=0.53]{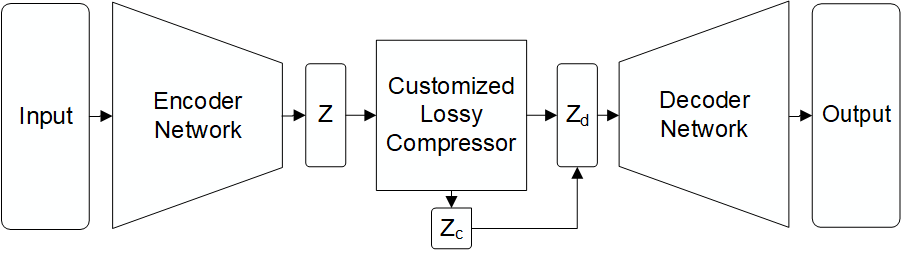}}
  \vspace{-1mm}
  \caption{The autoencoder with customized latent compressor}
  \label{fig:cae}
\end{figure}

One main disadvantage of autoencoders is the overhead of storing latent vectors, which can be reduced but cannot be eliminated. To maximize the compression ratio with autoencoders, instead of using the original encoder output latent vectors for compression, AE-SZ compresses the latent vectors with a built-in customized compressor and then uses the decompressed latent vectors for decoding. In this approach, the compressed latents are to be stored. The computation of compressed latents and autoencoder predictions in AE-SZ is shown in Figure \ref{fig:cae}. For the original latent vector $z$ as the encoder network, a lossy compressor generates the compressed latent $z_c$ in reduced size and the decompressed $z_d$ (which can also be directly computed from $z_c$); then the decoder network computes the prediction with $z_d$ as its input.

We customize an efficient method for compressing AE-SZ latent vectors (called \emph{customized} or \emph{custo.} for short) with two steps: (1) quantize the original value using an error bound of 0.1$e$, where $e$ is the user-specified error bound (the error bounds are value-range based) for the dataset; and (2) use Huffman + Zstd to compress the quantization codes. The advantage of such a design is twofold. First, this can get better compression ratios than SZ2.1, as shown in Table \ref{tab:basicandsz}. The key reason is that the latent vector data are not quite smooth across adjacent elements, based on our observation, while SZ2.1 strongly relies on spatial smoothness. Second, the custo. design is consistent with an important constraint required in the AE-SZ: the compression of each data block must be independent of other data blocks, which is explained as follows. Note that we select the better prediction method between AE and Lorenzo based on their prediction accuracy for each block. After this step, all the blocks (either AE-predicted blocks or Lorenzo-predicted blocks) have corresponding predicted data, which can be applied to the quantization directly. Obviously, in order to minimize the latent overhead, we should not store the AE latents for the Lorenzo-predicted blocks, but that requires that the compression of latents be independent across data blocks. SZ2.1 has data dependency across blocks, which makes it unsuitable for the latent vector compression here.  

\begin{table}[ht]
\centering
\footnotesize
  \caption {Compression ratios of our customized compressor vs. SZ2.1 on latent vectors under different error bounds $\epsilon$ } 
  \vspace{-2mm}
  \label{tab:basicandsz} 
  \begin{adjustbox}{width=\columnwidth}
  \begin{tabular}{|c|c|c|c|c|c|c|}
  \hline
  \multirow{2}{*}{\textbf{$\epsilon$}}&\multicolumn{2}{|c|}{\textbf{RTM}}&\multicolumn{2}{c|}{\textbf{NYX\_darkmatterdensity}}&\multicolumn{2}{c|}{\textbf{EXAFEL}}\\
    \cline{2-7}
    &\textbf{Custo.}&\textbf{SZ2.1}&\textbf{Custo.}&\textbf{SZ2.1}&\textbf{Custo.}&\textbf{SZ2.1}\\
    
    \hline
    1E$-$2 &\textbf{6.9}&5.9&\textbf{7.1}&6.2&\textbf{6.6}&5.7\\
    \hline
    1E$-$3 &\textbf{3.9}&3.4&\textbf{4.1}&3.6&\textbf{3.6}&2.9\\
    \hline
    1E$-$4 &\textbf{2.5}&2.0&\textbf{3.2}&2.5&\textbf{1.9}&1.4\\
    \hline
  
\end{tabular}
\end{adjustbox}
\end{table} 

\begin{figure}[ht]
  \centering
  \raisebox{-1mm}{\includegraphics[scale=0.5]{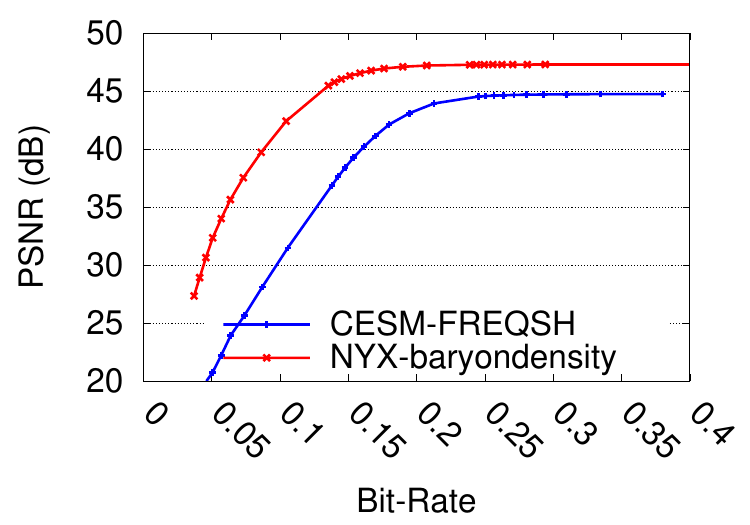}}
  \vspace{-1mm}
  \caption{Rate distortion of SWAE (without quantization) }
  \label{fig:ubaerd}
\end{figure}

Through masses of experiments using different datasets, we note that choosing a reasonable error bound can achieve a relatively high compression ratio of latents with a small loss of prediction accuracy. Figure \ref{fig:ubaerd} presents two rate-distortion plots of the AE prediction values with different compression ratios of latent vectors. The prediction accuracy (w.r.t. PSNR) does not degrade at all when the latent vectors are compressed with a ratio such as 4 (corresponding to bit-rate 0.25 in the figure as the original latent size is $\frac{1}{32}$ of the input size). That is, compressing latent vectors with a relatively high compression ratio (under a certain error bound) does not affect the compression of quantization bins much. 
%In conclusion, lossy compression of latent vectors brings considerable improvements to the final compression ratio. 
 
\subsection{Design Detail: Combination of AE and Lorenzo}
\label{sec:aepluslorenzo}

\noindent
\textbf{Takeaway 4}: \textit{The autoencoder model has a high ability to represent the data roughly with a high reduction ratio, but it is not as effective as Lorenzo in high-precision use cases. Therefore, a combination of AE and Lorenzo can effectively mitigate their own particular limitations in data prediction.}

Although AE has a great ability to learn the distribution of data, it still has two critical drawbacks that prevent it from being directly used as a data predictor, especially for high-precision error-bounded compression use cases. 
The first drawback is that, similar to linear regression, the latent vectors generated by AE for decompression sometimes bring cost due to redundancies of space. Specifically, we observe that quite a few data blocks may have constant or approximately constant values in scientific data. For these blocks, applying a  simple and low-cost predictor is accurate enough, while being able to reduce the storage size as much as possible. 
Second, to maintain learning effectiveness and efficiencies, the reconstructed data blocks from the autoencoder always suffer from certain noises, making it inadequate for extremely high-precision compression. By comparison, we note that the Lorenzo predictor outperforms the autoencoder especially when a relatively small error bound is used. 

%performs as a complement to autoencoders in 2-ways: For data blocks with simple patterns, it saves the space cost; For high precision (low error bound) compression, it provides more accurate predictions. 

\begin{figure}[ht] \centering

%\hspace{-10mm}
%\subfigure[{ErrBound=1E-1}]
%{
%\raisebox{-1cm}{\includegraphics[scale=0.38]{figures/prediction-error/FREQSH_55_1e-1_aggregated.predict.eps}}
%}
\hspace{-10mm}
\subfigure[{ErrBound=1E-2}]
{
\raisebox{-1cm}{\includegraphics[scale=0.38]{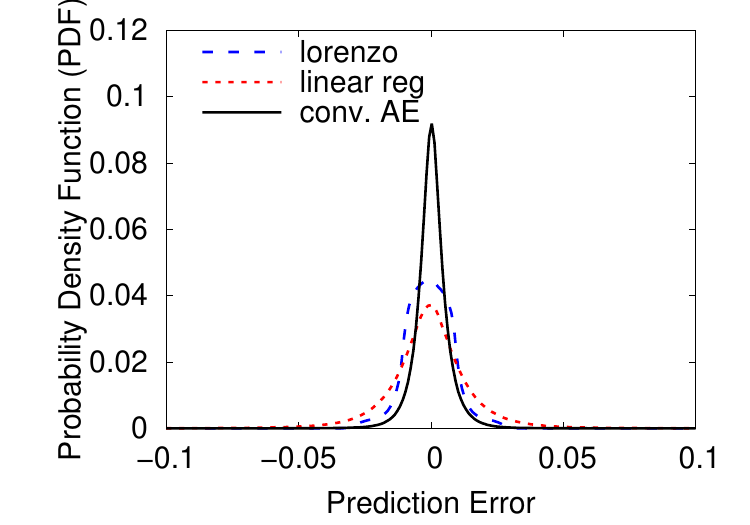}}
}
\hspace{-10mm}
%\hspace{-7mm}
%\subfigure[{ErrBound=1E-3}]
%{
%\raisebox{-1cm}{\includegraphics[scale=0.38]{figures/prediction-error/FREQSH_55_1e-3_aggregated.predict.eps}}
%}
%\hspace{-10mm}
\subfigure[{ErrBound=1E-4}]
{
\raisebox{-1cm}{\includegraphics[scale=0.38]{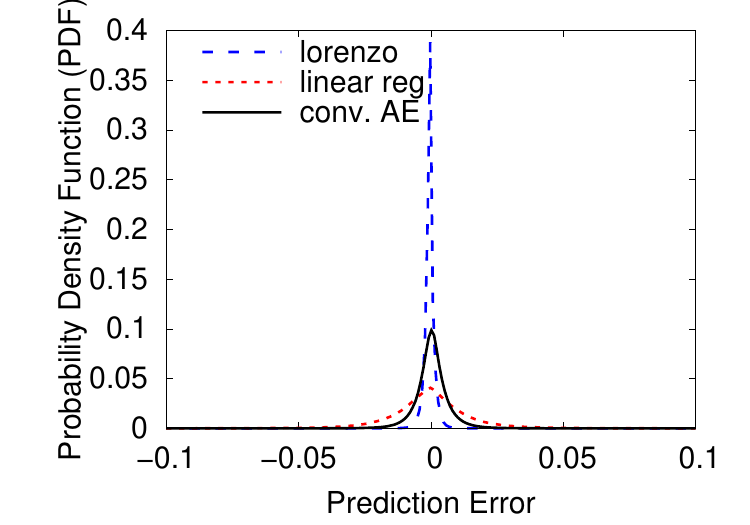}}
}
\hspace{-10mm}
\vspace{-3mm}

\caption{Distribution of prediction errors on CESM-FREQSH data field}
\label{fig:pred-err}
\end{figure}

We use Figure \ref{fig:pred-err} to illustrate the pros and cons of the autoencoder and Lorenzo predictor under different error bounds. This figure demonstrates the prediction error distributions of the Lorenzo predictor, linear regression predictor, and our trained autoencoder under an error bound of 1E-2 and 1E-4, respectively (the input data is a snapshot of the CESM-FREQSH data field). One can clearly observe that under the large error bound 1E-2, the autoencoder has a better (sharper) prediction error distribution. In contrast, the prediction accuracy of the Lorenzo predictor grows rapidly as the error bound decreases to a small value of 1E-4. %Therefore, we combine the two predictors in our compressor, and design an adaptive method to select the better one for each block during the compression.

During the online compression, AE-SZ selects a predictor between the autoencoder and Lorenzo for each data block. The selection criterion is checking which predictor has lower prediction errors (i.e., \textit{loss}) for the given block. The details can be found in Algorithm \ref{alg:aesz} (see line 6$\sim$13).

\section{Performance Evaluation}
\label{sec:evaluation}
%In this section, we first present the setups and configurations of the experiments on AE-SZ and other existing compressors for comparison, then provide the results together with our analysis and discussion of the results.

In this section, we present the experimental setup and then discuss the results.

\subsection{Experimental Setup}

\subsubsection{Experiment Environment}
We perform the experiments on the gpu\_v100\_smx2 nodes of the Argonne National Laboratory Joint Laboratory for System Evaluation computation cluster. Each node is driven by two Intel Xeon GOLD 6152 processors with 188 GB of DRAM and NVIDIA TESLA V100 GPUs. 
\subsubsection{Data Used in Experiments}

We perform the evaluation using five real-world application datasets in different domains that are commonly used in testing lossy compressors. Most of the datasets such as CESM, NYX, and Hurricane can be downloaded from SDRBench \cite{sdrb}. 
\begin{itemize}
    \item CESM \cite{cesm}: A well-known climate simulation package. We use its atmosphere model \cite{sdrb} in our experiments. These datasets are 2D, although some fields exhibit three dimensions in their metadata. For the CLOUD field (26$\times$1800$\times$3600), for instance, SZ2.1 has a better compression ratio (31.1 vs. 22.6) if we compress it with the range-based error bound 1E-3 in 2D mode instead of 3D mode.
    \item RTM: Reverse time migration (RTM) code for seismic imaging in areas with complex geological structures \cite{geodriveFirstBreak2020}.
    \item NYX \cite{nyx}: An adaptive mesh, hydrodynamics code designed to model astrophysical reacting flows on HPC systems. Two separate simulations are performed for the generation of training and test data. 
    \item Hurricane \cite{hurricane}: A simulation of a hurricane from the National Center for Atmospheric Research in the United States.
    \item EXAFEL \cite{exafel}: An Exascale Computing Project for analyzing molecular structure X-ray diffraction data generated by the LCLS \cite{lcls-ii}. The data contains groups of 32 2D arrays of size 185$\times$388. We discard the groups with nearly uniform data points; and following \cite{cappello2019use}, we concatenate the 2D arrays in each group to form a single 5920$\times$388 2D array for each group.
    
\end{itemize}

More detailed information on the datasets (all in single precision ) is shown in Table \ref{tab:dataset information}. The fields of NYX are transformed to their logarithmic value before compression for better visualization, as suggested by domain scientists. 

\begin{table}[ht]
    \vspace{-1mm}
    \centering
    \caption{Basic information about application datasets}  \vspace{-2mm}  
    %\footnotesize
\resizebox{0.99\columnwidth}{!}{        
    \begin{tabular}{|c|c|c|c|c|}
    \hline
    \textbf{App.}&\ \# \textbf{Files and fields}& \textbf{Dimensions} & \textbf{Fields used}& \textbf{Domain}\\
    \hline
    RTM &1 field, 3600 files&449$\times$449$\times$235&snapshot &Seismic Wave\\
    \hline
    \multirow{2}{*}{CESM} & \multirow{2}{*}{26 fields,62 files}& \multirow{2}{*}{1800$\times$3600}& CLDHGH& \multirow{2}{*}{Weather} \\
    & & &FREQSH& \\
    \hline
    EXAFEL & 1 field, 352 files & 5920$\times$388&raw data &Crystallography\\
    \hline
    NYX & 6 fields, 5 files & 512$\times$512$\times$512 & bd,t,dmd& Cosmology\\
    \hline
    Hurricane & 13 fields, 48 files & 100$\times$500$\times$500 &U, QVAPOR& Weather\\
    \hline
    \end{tabular}}
    \label{tab:dataset information}
\end{table}

\subsubsection{Comparison Lossy Compressors in Our Evaluation}

In our experiment, we compare AE-SZ with six other lossy compressors. The first four are classic error-bounded compressors: SZ2.1 \cite{sz16,sz17,Xin-bigdata18} and ZFP0.5.5 \cite{zfp}, which have been widely used in the community, and two recent works based on the SZ framework and developed from SZ2.1: SZauto \cite{Kai-HPDC2020} and SZinterp \cite{sz_interp}.  The fifth one is a recent work of an autoencoder-based scientific data compressor \cite{liu2021high}, called \textit{AE-A} in our evaluation. The sixth one is a pure convolutional autoencoder model \cite{glaws2020deep}, called \textit{AE-B}, proposed for compressing turbulence data, which is not error-bounded. 

%As \textit{AE-B} is not an error bounded compressor, its results are only shown in the rate distortion part (\ref{sec:evalrd}).
%The current versions of SZauto and SZinterp only support 3D data, so we only present results of these 2 compressors on 3D data fields.
%We make a comprehensive comparison with this compressor because it is a good representative of autoencoder-based scientific data compressor with available released codes.

\subsubsection{Experimental Configurations}
%In this section, we clarify the configurations of our experiments. 

For SZ2.1, ZFP0.5.5, SZauto, and SZinterp we adopt value-range-based error bounds and use default configurations for other parameters. 
%For the compression mode of compressors, in zfp we use fixed-accuracy mode (bounding error with given absolute error), and in all other compressor we use value-range based relative error bound mode. These are the mostly used compression modes in experiments of other works. Other parameters in SZ2.1 and zfp are default ones.

For the training phase of the autoencoders in AE-SZ, we train different autoencoders for different data fields on selected parts of the data, then test and compare all compressors on the remaining parts. Table \ref{tab:ae information} shows the input block size, length of latent vectors, number of the convolutional blocks in the encoder network, and number of channels in the convolutional blocks of the encoder network. The number of deconvolutional blocks is the same as the encoder's, and the channel numbers of the decoder network are symmetric with those in the encoder network. Table \ref{tab:split} shows the training-test split for all datasets. All autoencoders in AE-SZ are trained for 100 epochs.

%There are several more important details in the training of AE-SZ autoencoders:
%\begin{itemize}
%    \item As the train data of AE-SZ Autoencoder is also needed to be split into small blocks, we have found that in some datasets there are a great amount of approximately constant data blocks, and we remove these kind of blocks from training data. This is for 3 reasons: first, we observe that removing approximately constant data blocks from training can improve the prediction accuracy for the rest data blocks; Second, the approximately constant data blocks can be well predicted by Lorenzo predictor; Last, this process also improves the training speed as the training data size is reduced. Specifically speaking, we have removed data blocks from training which have relative ranges (range divides global range) under 1e-4.
%    \item For each data field, when doing training and prediction, the input of Autoencoder is linearly normalized to the range of [-1, 1] based on the global maximum and minimum of data, and the output is denormalized in reverse to get the final predictions. 
%\end{itemize}

%\begin{comment}
\begin{table}[ht]
    \vspace{-1mm}
    \centering
    \caption{Autoencoder Configurations for Each data Field}  \vspace{-2mm}  
    %\footnotesize
\resizebox{0.99\columnwidth}{!}{        
    \begin{tabular}{|c|c|c|c|c|}
    \hline
    \textbf{Data field}&\ \textbf{Input block}& \textbf{Latent size} & \textbf{Block num.}& Channels\\
    \hline
    CESM-CLDHGH &32x32&16&4 &[32,64,128,256]\\
    \hline
    CESM-FREQSH& 32x32& 32& 4& [32,64,128,256] \\
    \hline
    EXAFEL & 32x32 & 16& 4 &[32,64,128,256]\\
    \hline
    RTM & 16x16x16 & 16 & 4 &[32,64,128,256]\\
    \hline
    NYX (all fields) & 8x8x8 & 16 & 3 & [32,64,128]\\
    \hline
    Hurricane-U & 8x8x8 & 8 & 3 & [32,64,128]\\
    \hline
    Hurricane-QVAPOR & 8x8x8 & 16 & 3 & [32,64,128]\\
    \hline

    \end{tabular}}
    \label{tab:ae information}
\end{table}
%\end{comment}

\begin{table}[ht]
    \vspace{-1mm}
    \centering
    \caption{Train-test split for each dataset}  \vspace{-2mm}  
    \footnotesize
%\resizebox{0.99\columnwidth}{!}{        
    \begin{tabular}{|c|c|c|}
    \hline
    \textbf{Dataset}&\ \textbf{Train split}& \textbf{Test split}\\
    \hline
    CESM &[0,49]&[50,62]\\
    
    \hline
    EXAFEL & [0,299] & [300,351]\\
    \hline
    RTM & [1400,1499] & 1510 to 1600 step 10\\
    \hline
    NYX  & redshift [54,42] & another simulation at redshift 42\\
    \hline
    Hurricane & [1,40] & [41,48]\\
    \hline
    \end{tabular}
%    }
    \label{tab:split}
\end{table}

For \textit{AE-A} \cite{liu2021high}, we download their from https://github.com/tobivcu/autoencoder, which supports only double-precision floating data originally. We improve the code by enabling it to compress single-precision floating data, in that most of the datasets in our test are stored in single-precision. We trained its model using the same training data split for 100 epochs. The .dvalue files generated by the model are compressed by SZ2.1 following the instruction of \cite{liu2021high}. After fine-tuning, we applied the same value-range-based relative error bound to compress the .dvalue file. 
%For the compression of .dvalue files generated by their model, [] presents 2 ways: saved it in half precision type, or compress them with SZ2.1.%

%These 2 methods of dealing with .dvalue files are named as []\_fp16 and []\_sz in later result presentation parts. For the compression ratio-error bound part, we just display the best compress ratios among these two configurations.    

For \textit{AE-B}, since  Glaws et al. \cite{glaws2020deep} does not provide enough details for training from scratch, following the paper's recommendation, we fine-tuned a pretrained autoencoder (from https://github.com/NREL/AEflow indicated by \cite{glaws2020deep}) on different data fields for 5 epochs each. 
%Note that we only test \textit{AE-B} on RTM, NYX and Hurricane data since their model is currently only designed for 3D input data.

\subsubsection{Evaluation Metrics}

We evaluate the seven lossy compressors based on the critical metrics described below. 
\begin{itemize}
    %\item Compression ratio (CR) based on the same error bound: The descriptions of CR and absolute error bound are defined in Section \ref{sec:problem}.
    \item \textit{Rate distortion}: Rate distortion is the most commonly used metric by the lossy compression community to assess compression quality. Rate distortion involves two critical metrics: peak signal-to-noise ratio (PSNR) and bit rate. The definition of PSNR is introduced in section \ref{sec:problem}, and bit rate is defined as the average number of bits used per data point in the compressed data. Generally speaking, Bit rate equals $Sizeof(datatype)/cr$, in which $Sizeof(datatype)$ is the byte size of input data (32 for single-precision data for example), and $cr$ is the compression ratio. Therefore, a smaller bit rate means a better compression ratio, and vice versa. 
    \item Visualization with the same compression ratio (CR): Compare the visual quality of the reconstructed data based on the same CR.    
    \item Compression speed and decompression speed: $\frac{original \hspace{0.5mm} size}{compression\hspace{0.5mm} time}$ (MB/s) and $\frac{reconstructed\hspace{0.5mm} size}{decompression\hspace{0.5mm} time}$ (MB/s).
    In the following experimental results, when it comes to error bound values, without loss of generality, we adopt value-range-based error bounds (denoted as $\epsilon$), which takes the same effect with absolute error bound (denoted $e$) because $e=\epsilon\cdot(\max(D)-\min(D))$. 
\end{itemize}

\subsection{Evaluation Results and Analysis}

\subsubsection{Rate distortions of different lossy compressors}
\label{sec:evalrd}
\begin{figure}[ht] \centering

\hspace{-17mm}
\subfigure[{CESM-CLDHGH (2D)}]
{
\raisebox{-1cm}{\includegraphics[scale=0.4]{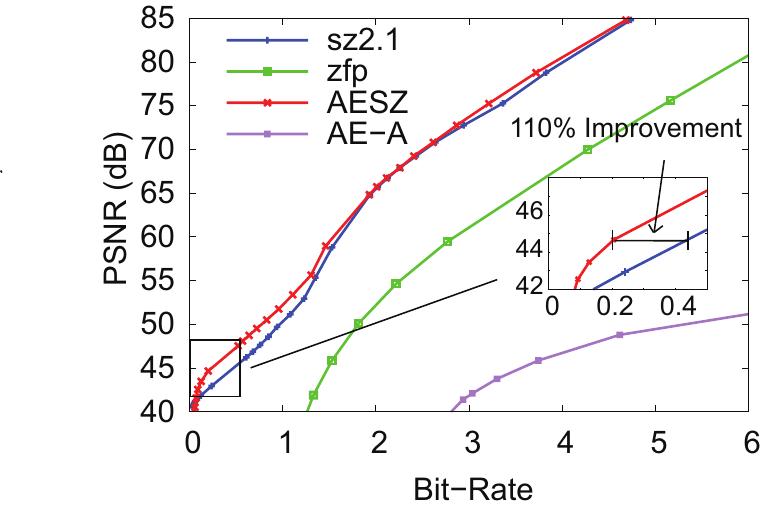}}
}
\hspace{-7mm}
\subfigure[{CESM-FREQSH (2D)}]
{
\raisebox{-1cm}{\includegraphics[scale=0.4]{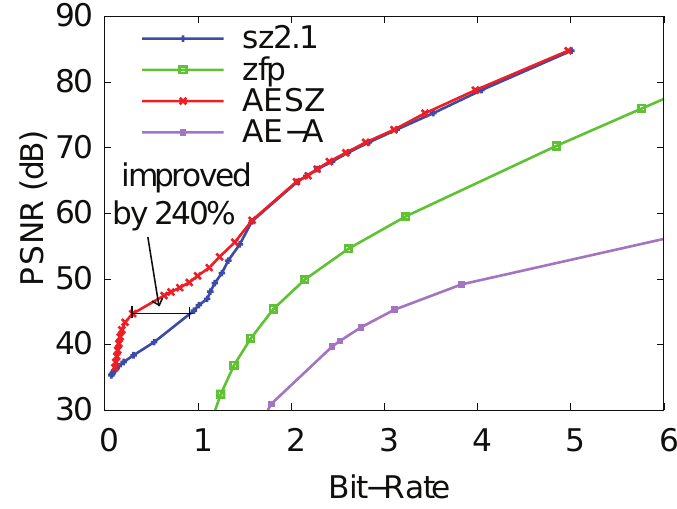}}
}
\hspace{-10mm}

\hspace{-11mm}
\subfigure[{EXAFEL (2D)}]
{
\raisebox{-1cm}{\includegraphics[scale=0.4]{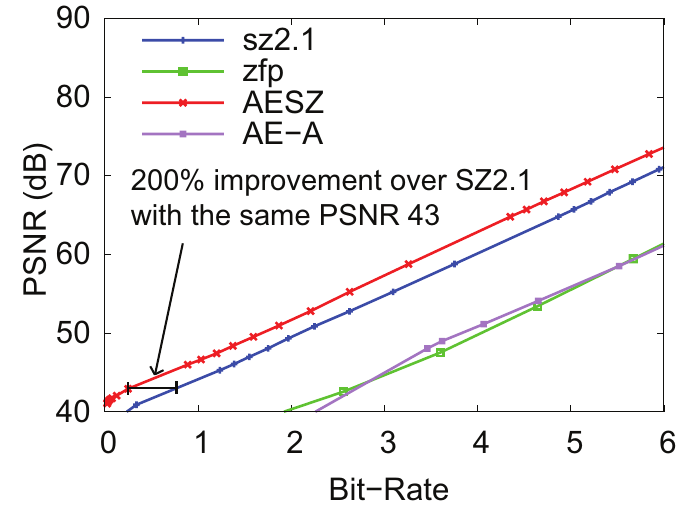}}
}
\hspace{-7mm}
\subfigure[{NYX-baryon\_density (3D)}]
{
\raisebox{-1cm}{\includegraphics[scale=0.4]{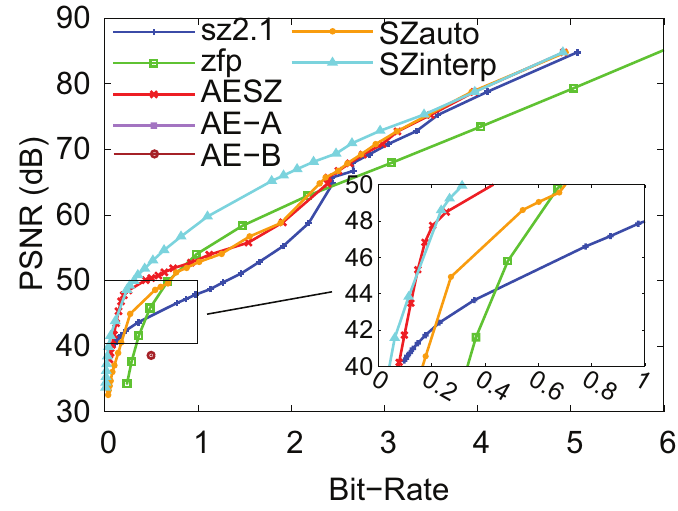}}
}
\hspace{-10mm}

\hspace{-11mm}
\subfigure[{NYX-temperature (3D)}]
{
\raisebox{-1cm}{\includegraphics[scale=0.4]{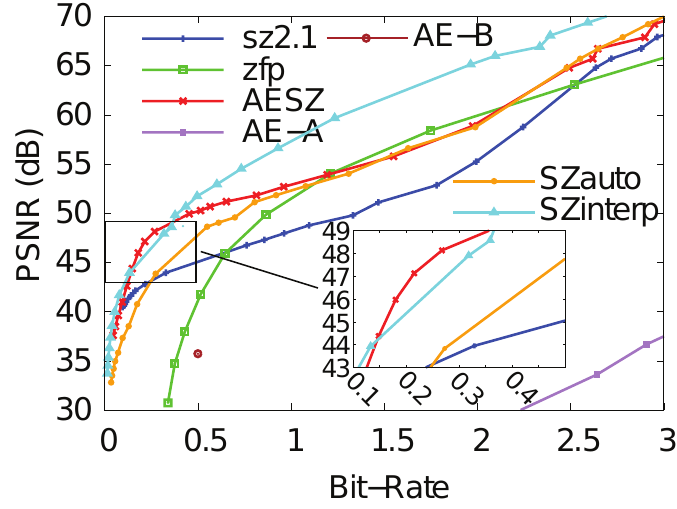}}
}
\hspace{-7mm}
\subfigure[{Hurricane-QVAPOR (3D)}]
{
\raisebox{-1cm}{\includegraphics[scale=0.4]{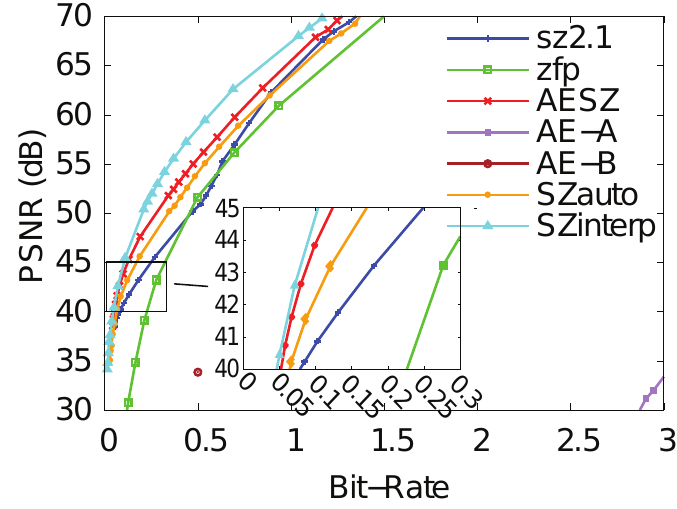}}
}
\hspace{-10mm}

\hspace{-10mm}
\subfigure[{Hurricane-U (3D)}]
{
\raisebox{-1cm}{\includegraphics[scale=0.4]{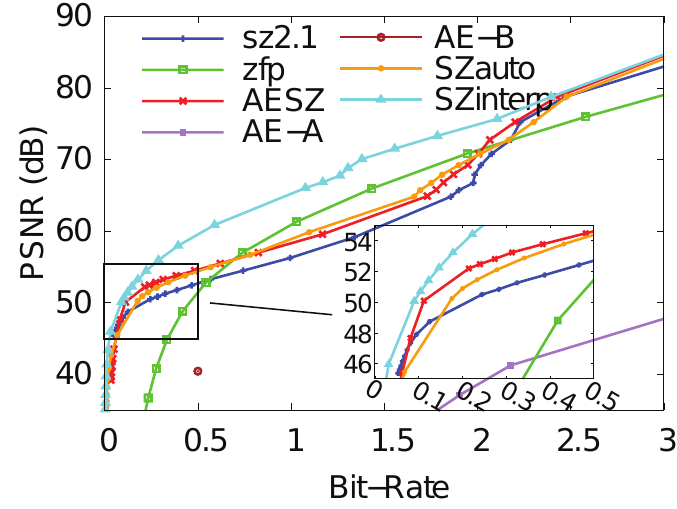}}
}
\hspace{-7mm}
\subfigure[{RTM (3D)}]
{
\raisebox{-1cm}{\includegraphics[scale=0.4]{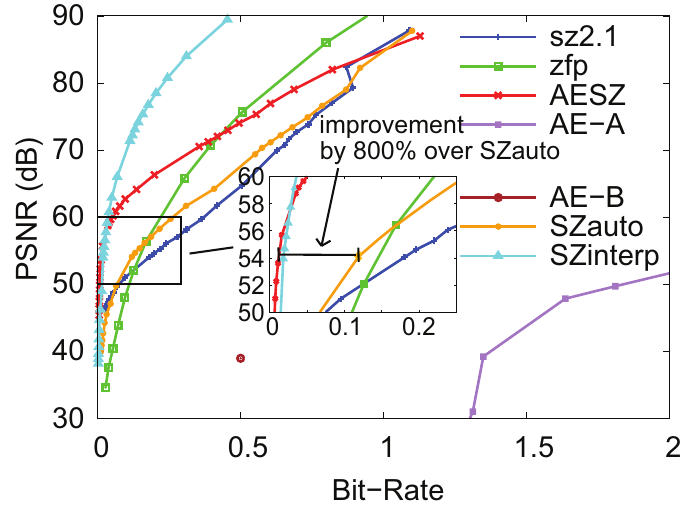}}
}
\hspace{-10mm}
\vspace{-3mm}

\caption{Rate distortion of different compressors}
\label{fig:rate-distortion}
\end{figure}

We present the rate-distortion results of all seven lossy compressors on all tested data fields, illustrating the PSNR of final decompression results with bit rates. 
%Bit rate indicates the average number of bits used to represent one data value, so the smaller bit-rate the larger compression ratio. 
%We test each compressor under various error bounds, and plot rate distortion curves. 
Figure \ref{fig:rate-distortion} shows the rate-distortion plots for each lossy compressor on eight data fields. Only four compressors are shown in Figure \ref{fig:rate-distortion} (a), (b), and (c) because the other three compressors (SZauto, SZinterp, and \textit{AE-B}) support only 3D data, while CESM and EXAFEL are both 2D datasets. We observe that AE-SZ is significantly better than the other two AE-based lossy compressors (\textit{AE-A} and \textit{AE-B}) in terms of rate distortion. That is, our developed AE-SZ compression method is arguably the best AE-based lossy compressor to date. We also compare the most competitive error-bounded lossy compressors (to the best of our knowledge):  SZinterp \cite{sz_interp}, SZauto \cite{Kai-HPDC2020}, SZ2.1 \cite{Xin-bigdata18}, and ZFP \cite{zfp}. Generally speaking, AE-SZ obtains much better rate distortions than SZauto, SZ2.1, and ZFP do under low bit rates (i.e., in high-compression-ratio cases) and have a comparable quality with SZ2.1 for high bit rates. We observe that AE-SZ generally has 100\%$\sim$800\% higher compression ratios than SZ2.1 has in the high-compression-ratio cases on both 2D and 3D datasets. In the 2D datasets, for example, AE-SZ exhibits the best rate distortion (240\% higher compression ratio than the second best at the same PSNR around 44) for the CESM-FREQSH data field. On the EXAFEL dataset, AE-SZ has a 200\% higher compression ratio than the second best (SZ2.1) in the high-compression cases. On the 3D datasets, AE-SZ also exhibits very competitive rate distortions among all the seven compressors. Its compression quality is close to that of SZinterp in the low-bit-rate range (e.g., [0,1]) and may also exhibit the best rate-distortion in a few cases (e.g., Figure \ref{fig:rate-distortion} (e)). 

%In conclusion, in the metric of rate-distortion, compared with other tested lossy compressors, AE-SZ is the best one under low bit rates and is at least as good as SZ2.1 on high bit rates. On medium bit rates, AE-SZ works best on 2D data fields and partially best on 3D data fields, with zfp sometimes over-performing it in the rate-distortion scale under medium bit rates on parts of 3D data fields. As zfp suffers from high maximum point-wise decompression errors (refer to table \ref{tab:compress ratio comparison}, zfp always get lower compression ratios on fixed error bounds), in these cases AE-SZ can still make advantages over zfp. 
\subsubsection{Decompression data visualizations of different lossy compressors}
\begin{figure}[ht] \centering

\hspace{-8mm}
\subfigure[{Original}]
{
\raisebox{-1cm}{\includegraphics[scale=0.2]{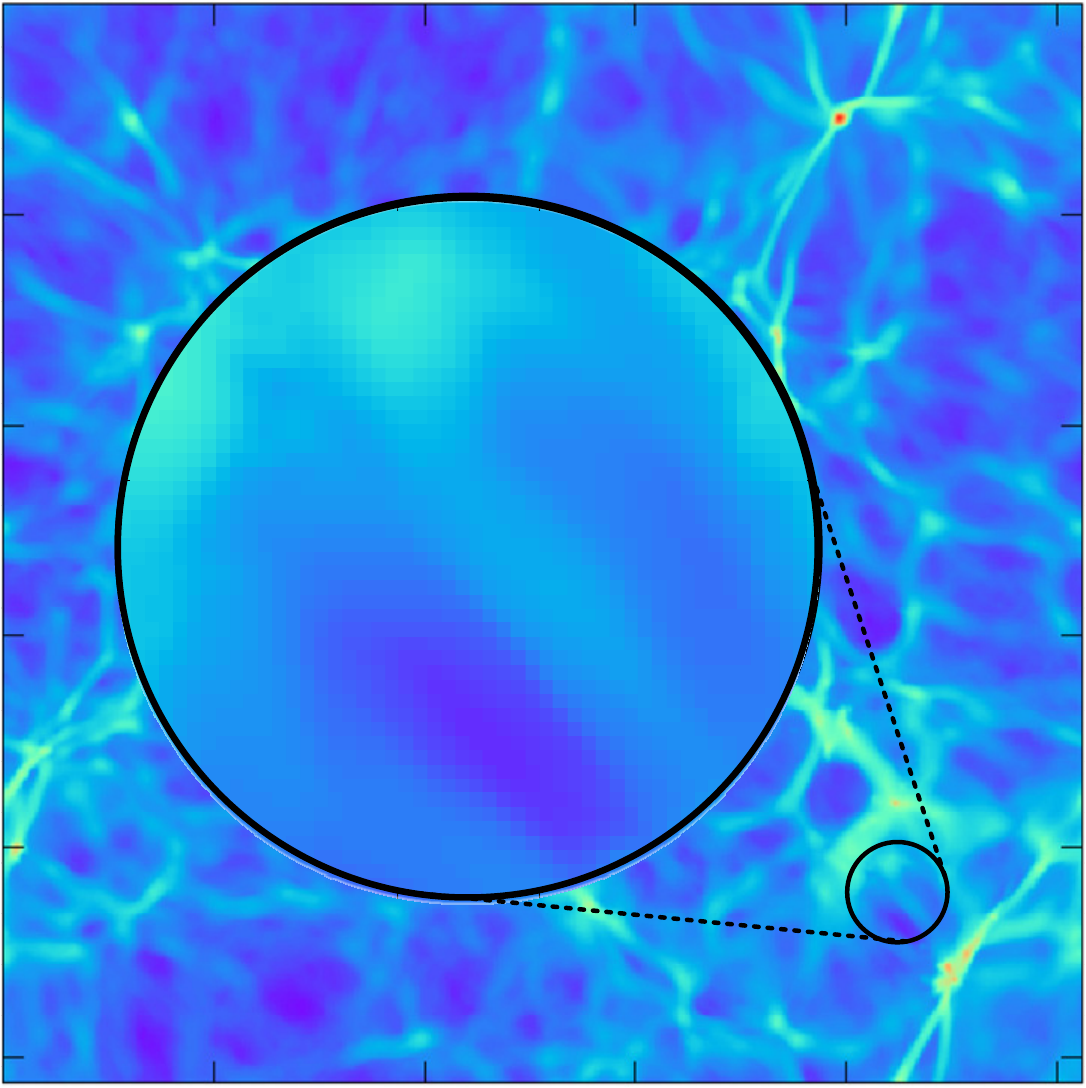}}
}
\hspace{-3mm}
\subfigure[{AE-SZ (PSNR:46.8,CR:182.3)}]
{
\raisebox{-1cm}{\includegraphics[scale=0.2]{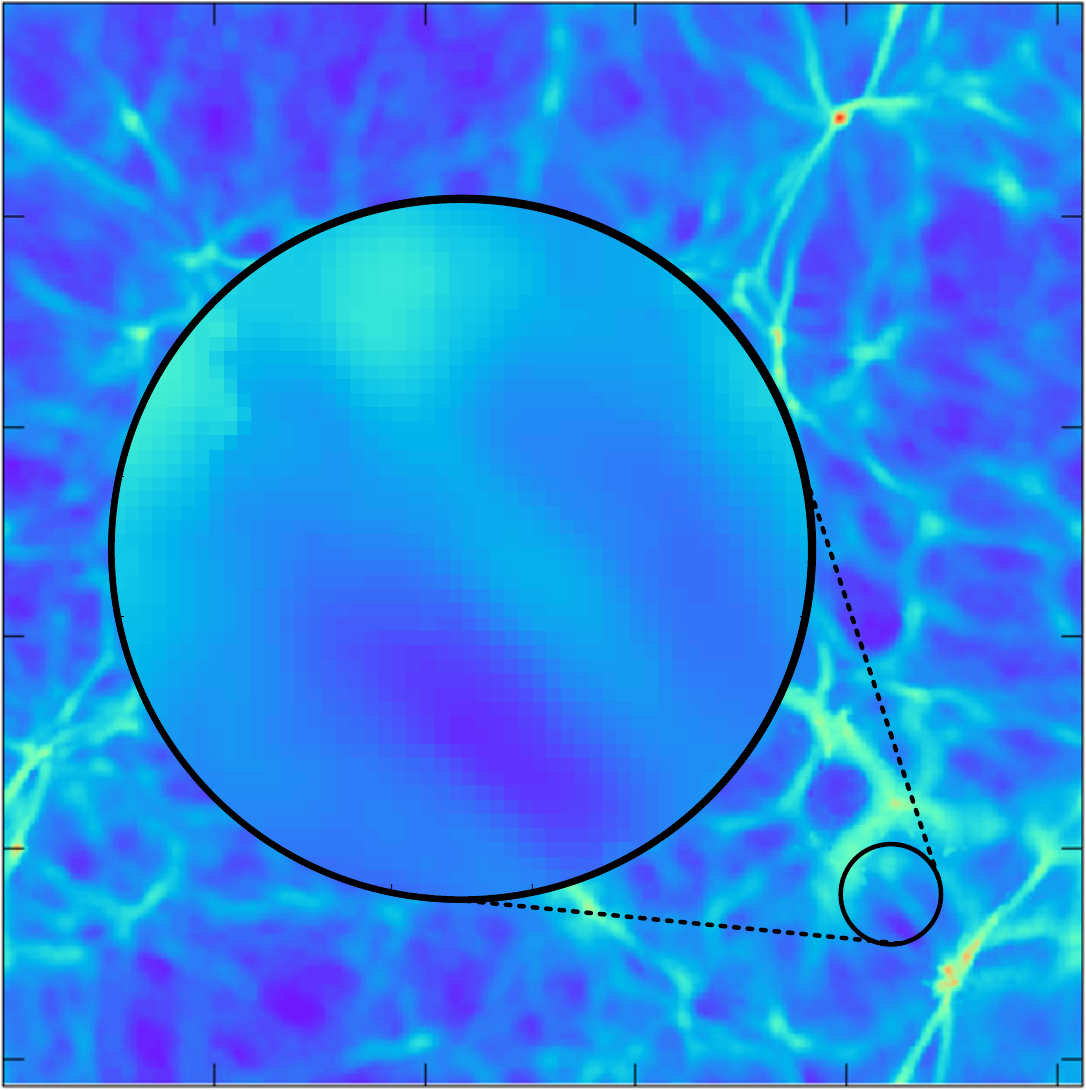}}
}
\hspace{-10mm}

\hspace{-7mm}
\subfigure[{SZinterp (PSNR:45.5,CR:182)}]
{
\raisebox{-1cm}{\includegraphics[scale=0.2]{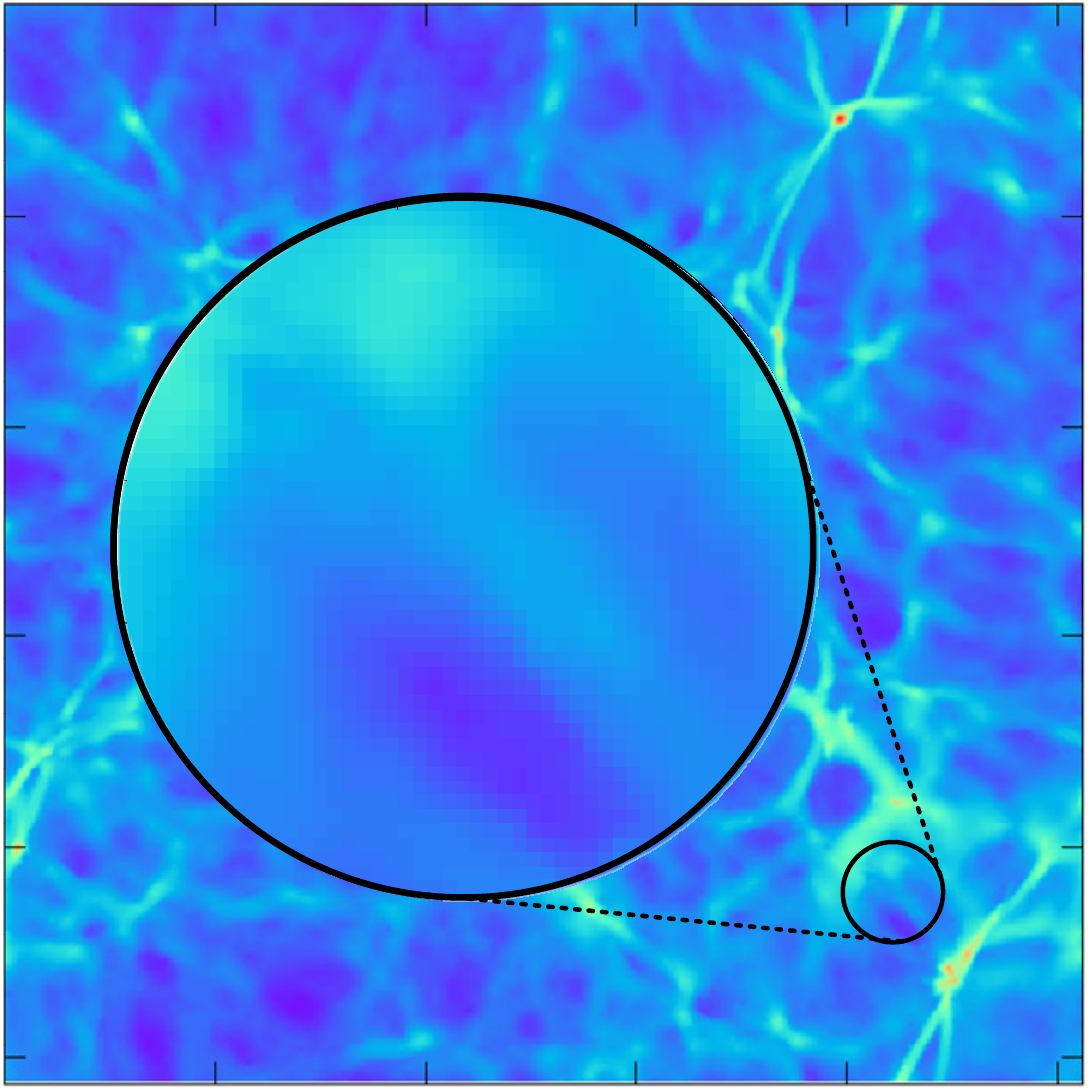}}
}
\hspace{-3mm}
\subfigure[{SZauto (PSNR:40.6,CR:179)}]
{
\raisebox{-1cm}{\includegraphics[scale=0.2]{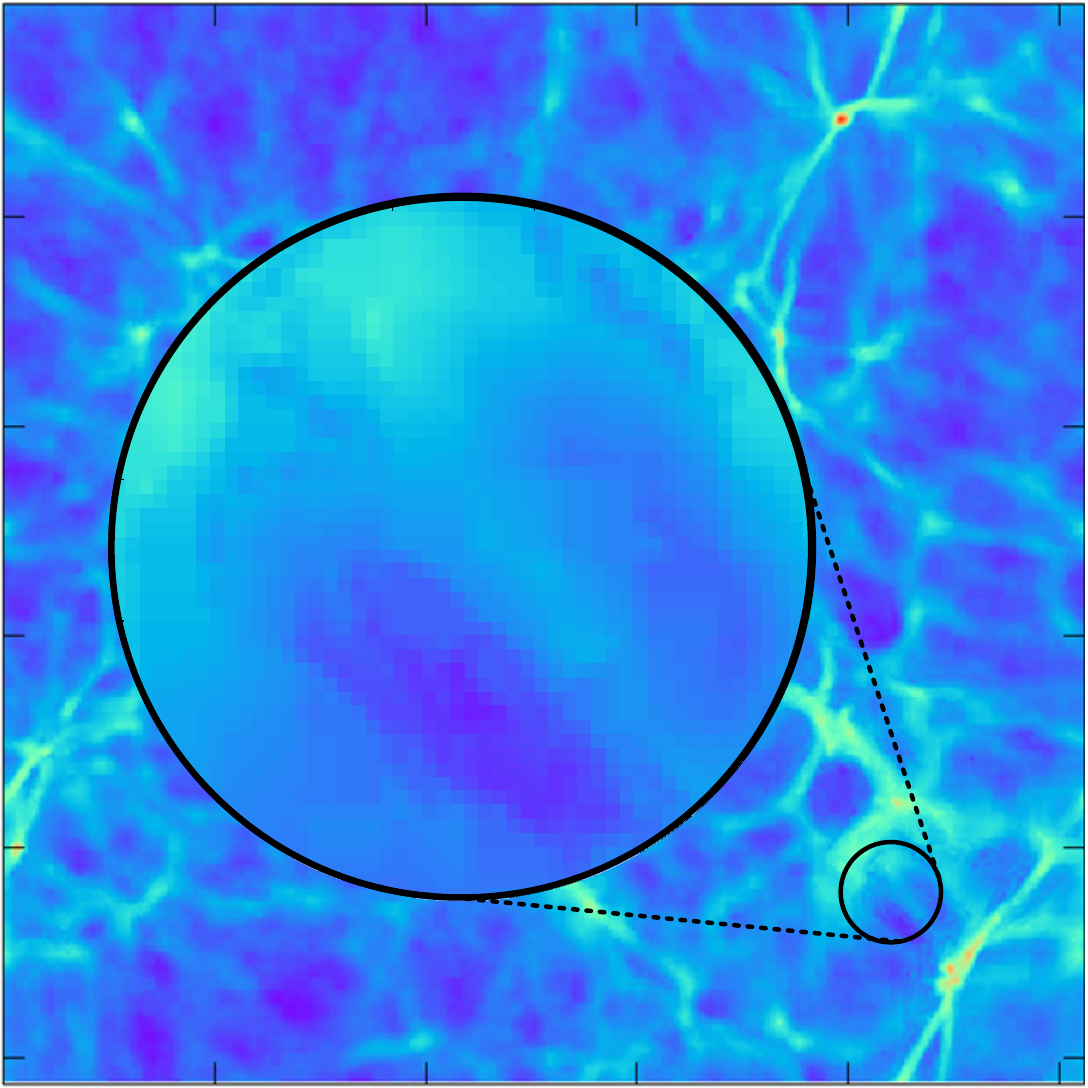}}
}
\hspace{-10mm}

\hspace{-7mm}
\subfigure[{SZ (PSNR:41.7,CR:182.5)}]
{
\raisebox{-1cm}{\includegraphics[scale=0.2]{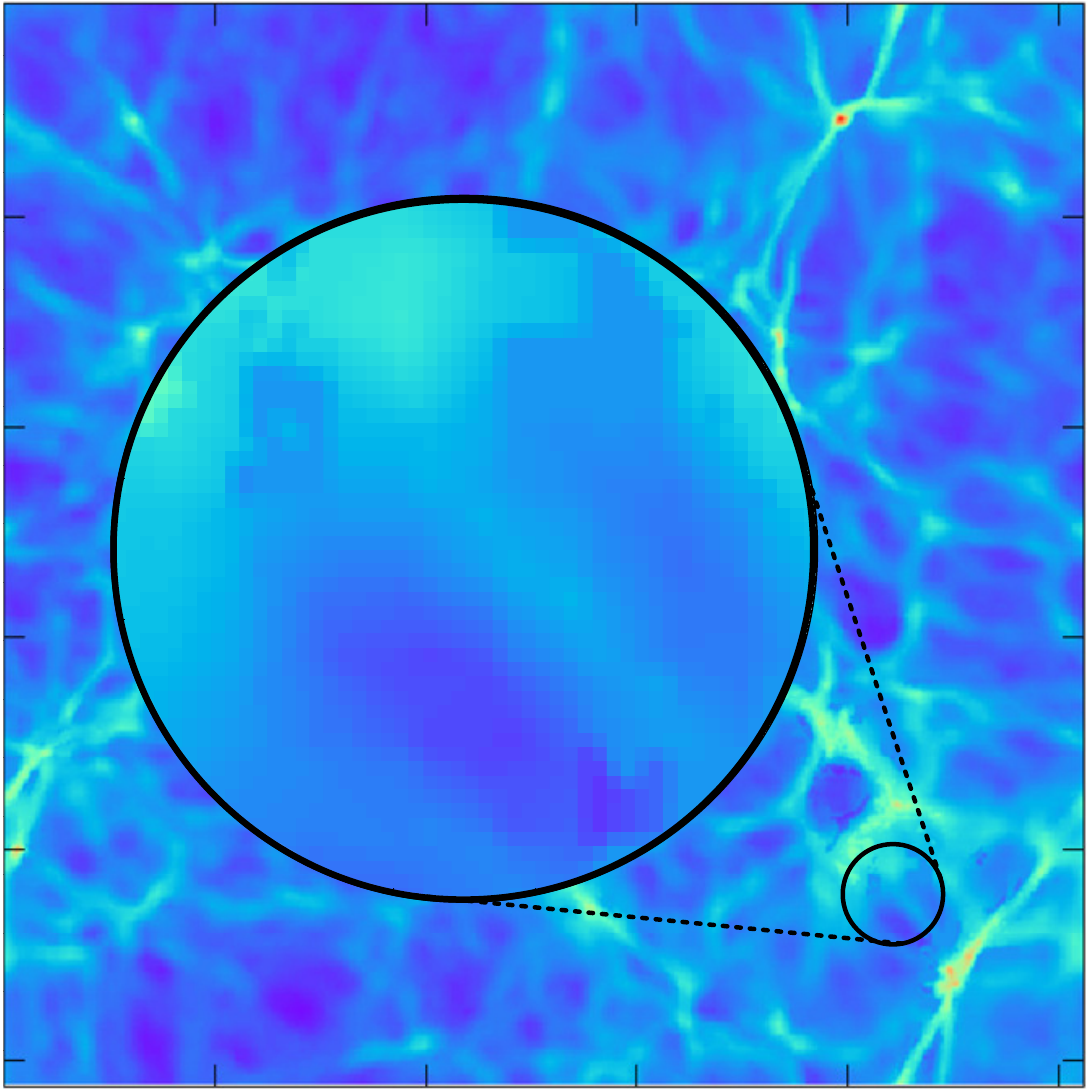}}
}
\hspace{-3mm}
\subfigure[{ZFP (PSNR:30.2,CR:161)}]
{
\raisebox{-1cm}{\includegraphics[scale=0.2]{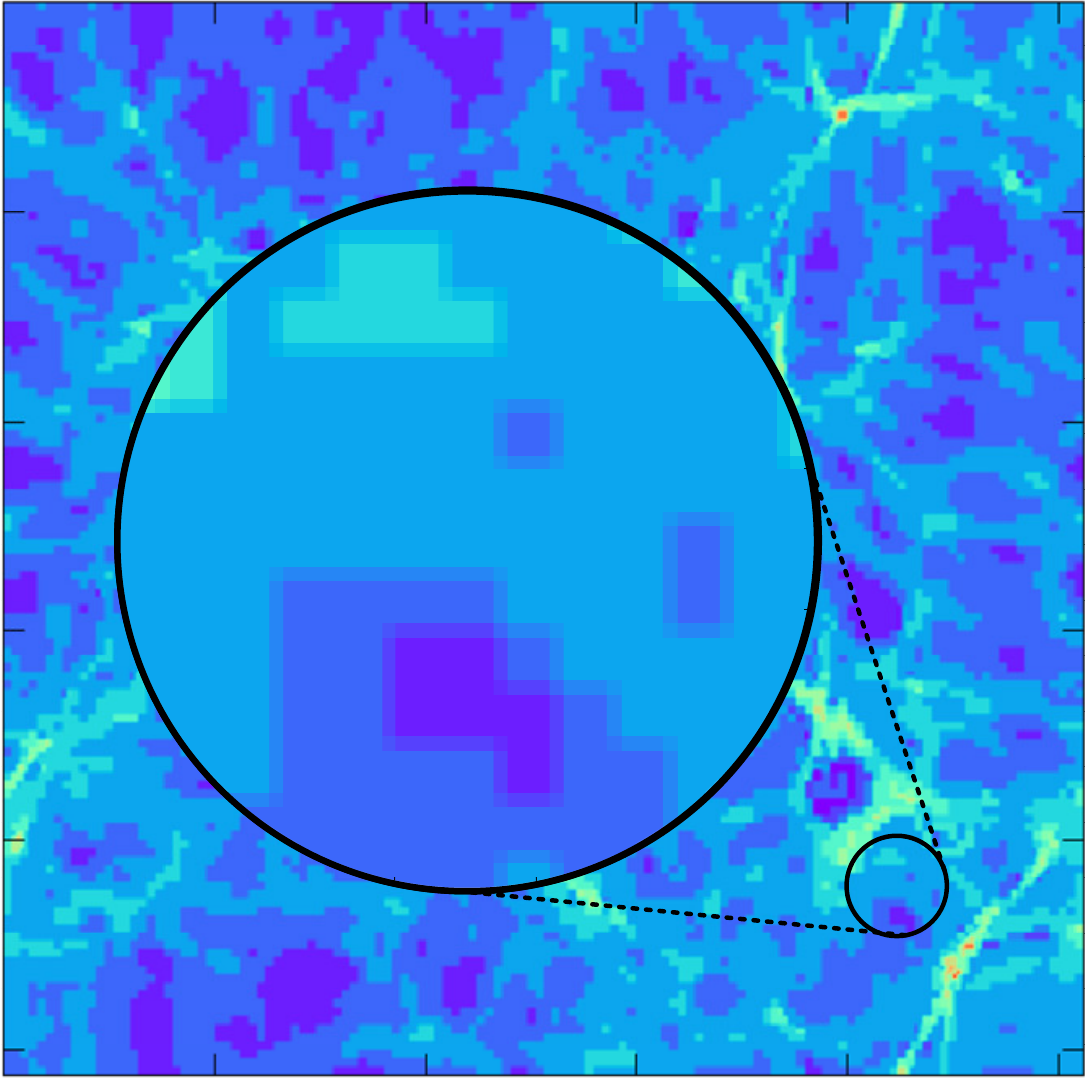}}
}
\hspace{-10mm}
\vspace{-3mm}

\caption{Visualization of reconstructed data (NYX-baryon\_density)}
\label{fig:vis-temp}
\end{figure}

We present data visualizations in Figure \ref{fig:vis-temp} on the NYX-baryon\_density field to verify the effectiveness of the reconstructed data of AE-SZ at high compression ratio use cases. We clearly observe that the reconstructed data at the PSNR of 46.8 under AE-SZ has a very good visual quality. Other prior works \cite{Xin-bigdata18,liang2019significantly} show that PSNR in the range of [30,60] is good enough to have a high visual quality for different scientific applications. 
Moreover, Figure \ref{fig:vis-temp} demonstrates that with the same compression ratio of 180, AE-SZ has a much better visual quality compared with that of the three state-of-the-art lossy compressors, SZauto, SZ2.1, and ZFP0.5, and is also better than SZinterp. 

\subsubsection{Performances of AE-SZ predictors under different error bounds}
\begin{figure}[ht]
  \centering
  \raisebox{-1mm}{\includegraphics[scale=0.5]{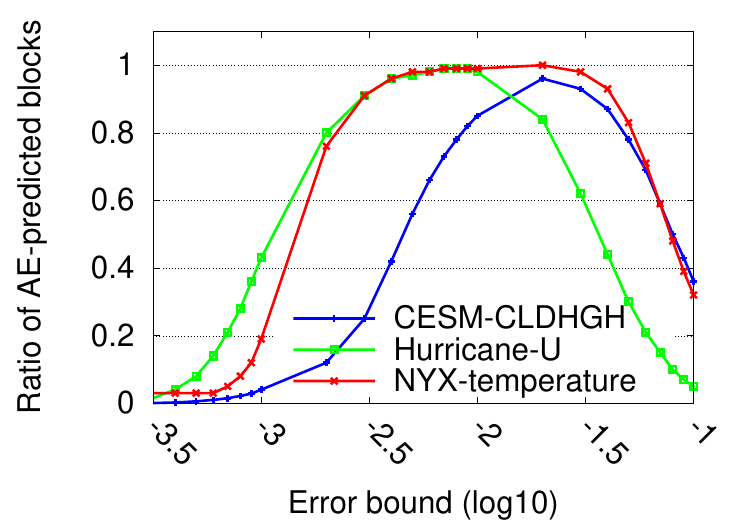}}
  \vspace{-1mm}
  \caption{Percentage of blocks predicted by AE under different error bounds}
  \label{fig:nnratio}
\end{figure}

To better understand how the autoencoder and Lorenzo predictor in AE-SZ cooperatively contribute to the compression ratios, we record the percentage of data blocks predicted by AE-SZ autoencoders on three different data fields, as shown in Figure \ref{fig:nnratio}.
For better vision of the plots, the x-axis is logged error bounds. The plots show that autoencoders in AE-SZ achieve advantages over Lorenzo under a range of medium error bounds (about 5E-3 to 2E-2), under which most of the data blocks can be better predicted by autoencoders. As the error bound decreases, the Lorenzo predictor becomes better than autoencoders on more data blocks. When the error bound becomes very high, the latents need to be compressed with a high error bound, so the prediction error of autoencoders may drop rapidly, and Lorenzo may also turn better.

\begin{figure}[ht] \centering
%\hspace{-10mm}
\subfigure[{CESM (CLDHGH)}]
{
\raisebox{-1cm}{\includegraphics[scale=0.45]{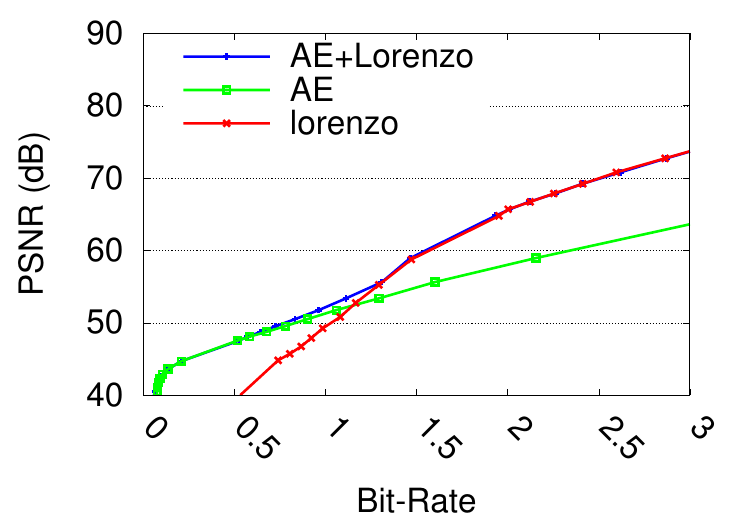}}
}
%\hspace{-7mm}
\subfigure[{Hurricane (U)}]
{
\raisebox{-1cm}{\includegraphics[scale=0.45]{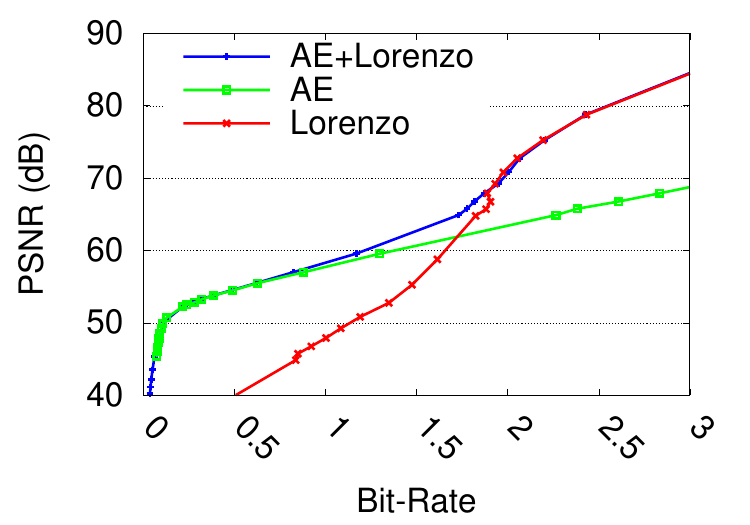}}
}
%\hspace{-10mm}
\vspace{-1mm}

\caption{AE-SZ rate distortion comparison of predicting data by AE+Lorenzo, AE only, or Lorenzo only}
\label{fig:ratedist-breakdown}
\end{figure}

To understand the effectiveness of our adaptive prediction design, we present the rate-distortion in three situations: predicting data with only AE, predicting data with only Lorenzo, and combining both, as shown in Figure \ref{fig:ratedist-breakdown}. The figure shows that AE+Lorenzo achieves the best quality at all bit rates since it can take advantage of both predictors adaptively. 

%\vspace{-1mm}

%\vspace{11mm}
%\vspace{3mm}

\subsubsection{Compression speeds and autoencoder training speeds}

The average compression speed of each error-bounded lossy compressor on all the tested datasets under the error bound of 1E-3 are shown in Table \ref{tab:compressionspeed} in units of Mb/s (SZauto, SZinterp, and \textit{AE-B}  have speeds only on 3D data because they currently do not support 2D data). Because of the relatively high computation cost of neural networks, AE-SZ cannot achieve comparable compression throughput with traditional lossy compressors (its speed is about 10\%-40\% as fast as that of SZ2.1 and SZinterp). In fact, the current version of the AE-SZ code is in the experimental stage, so it is not as optimized as the off-the-shelf compressors such as SZ2.1 and ZFP. We believe that with further optimization AE-SZ can be much accelerated. In fact, the throughput of AE-SZ has been significantly better than the other autoencoder-based compressors such as \textit{AE-A} \cite{liu2021high} by 30$\times$ to 200$\times$ (mainly due to the complicated data preprocessing and postprocessing procedures in \textit{AE-A}) and \textit{AE-B} \cite{glaws2020deep} by up to 4$\times$ speedup in compression and 9$\times$ speedup in decompression (note that \textit{AE-B} is not error bounded so its running speeds include only the AE prediction process). 
%Higher compression ratios could be more important than higher compression speeds in many use-cases.
%Also note that AE-SZ best performance is for low bit rate (high compression ratio) compression, and in this kind of task having higher compression speed is often less important than achieving high compression ratio. 

\begin{table}[ht]
\centering
\footnotesize
  \caption {Compression/decompression speeds (MB/s): error bound=1E-3} 
  \vspace{-2mm}
  \label{tab:compressionspeed} 
  \begin{adjustbox}{width=\columnwidth}
  \begin{tabular}{|c|c|c|c|c|c|c|c|c|}
  \hline
      \multirow{2}{*}{\textbf{Type}} & \multirow{2}{*}{\textbf{Dataset}} & \textbf{SZ}  & \textbf{ZFP} & \textbf{SZ}& \textbf{SZ} & \multirow{2}{*}{\textbf{AE-SZ}}  & \multirow{2}{*}{\textbf{AE-A\cite{liu2021high}}} & \multirow{2}{*}{\textbf{AE-B\cite{glaws2020deep}}} \\ &  & \textbf{2.1}  & \textbf{0.5}& \textbf{auto}& \textbf{interp}&  & &  \\ \hline
    \multirow{5}{*}{\rotatebox[origin=c]{90}{Comp}} & CESM & 145.0 & 174.7 &N/A&N/A& 26.7 & 0.4&N/A    \\ \cline{2-9}
    & RTM & 196.4 & 432.4 & 231.4 & 183.2 & 68.2 & 0.4 & 15.6   \\ \cline{2-9}
    & Hurricane & 166.2 & 91.8 & 198.7 & 168.6 & 22.0 & 0.4& 15.5    \\ \cline{2-9}
    & NYX & 142.9 & 171.7 & 152.9 & 99.6 & 15.7 & 0.5 & 16.0 \\ \cline{2-9}
    & EXAFEL & 162.3 & 213.7 & N/A & N/A & 12.2 & 0.4 & N/A  \\ \cline{1-9}
    \multirow{5}{*}{\rotatebox[origin=c]{90}{Decomp}} & CESM & 249.7 & 222.7 & N/A & N/A & 58.0 & 0.6 & N/A   \\ \cline{2-9}
    & RTM & 430.3 & 941.3 & 516.1 & 452.6 & 135.3 & 0.7 & 14.1    \\ \cline{2-9}
    & Hurricane & 264.9 & 243.1 & 328.6 & 357.4 & 48.2 & 0.6 & 14.7  \\ \cline{2-9}
    & NYX & 217.7 & 310.4 & 197.9 & 115.6 & 34.6 & 0.7 & 14.4 \\ \cline{2-9}
    & EXAFEL & 219.1 & 224.7 & N/A & N/A & 26.1 & 0.6 & N/A  \\ \cline{1-9}
\end{tabular}
\end{adjustbox}
\end{table}

Second, table \ref{tab:trainingspeed} shows the training time of autoencoders in AE-SZ and \textit{AE-A} \cite{liu2021high} using the same training data and the same number of training epochs. We can conclude that the autoencoders in AE-SZ outperform \textit{AE-A} \cite{liu2021high} with similar or shorter training time. For \textit{AE-B}, the tested networks are only fine-tuned so we are unable to present its training time.

\begin{table}[ht]
\centering
\footnotesize
  \caption {Autoencoder Training Time (in hours) } 
  \vspace{-2mm}
  \label{tab:trainingspeed} 
  %\begin{adjustbox}{width=\columnwidth}
  \begin{tabular}{|c|c|c|}
  \hline
    \textbf{Dataset}&\ \textbf{AE-SZ}& \textbf{AE-A \cite{liu2021high}}\\
    \hline
    CESM &1.0&1.5\\
    \hline
    RTM & 3.4 & 21.4\\
    \hline
    NYX  & 5.5 & 4.7\\
    \hline
    Hurricane & 2.4 & 2.5\\
    \hline
    EXAFEL & 2.2 & 3.5\\
    \hline
\end{tabular}
%\end{adjustbox}
\end{table}

%\begin{figure}[ht] \centering
%
%\hspace{-8mm}
%\subfigure[{Original}]
%{
%\raisebox{-1cm}{\includegraphics[scale=0.23]{figures/visualization/FREQSH_61_ori.eps}}
%}
%\hspace{-3mm}
%\subfigure[{AE-SZ (PSNR:43.7,CR:99.4)}]
%{
%\raisebox{-1cm}{\includegraphics[scale=0.23]{figures/visualization/FREQSH_61_aesz.eps}}
%}
%\hspace{-10mm}
%
%\hspace{-7mm}
%\subfigure[{SZ (PSNR:38.3,CR:98.9)}]
%{
%\raisebox{-1cm}{\includegraphics[scale=0.23]{figures/visualization/FREQSH_61_sz.eps}}
%}
%\hspace{-3mm}
%\subfigure[{ZFP (PSNR:22.3,CR:34)}]
%{
%\raisebox{-1cm}{\includegraphics[scale=0.23]{figures/visualization/FREQSH_61_zfp.eps}}
%}
%\hspace{-10mm}
%\vspace{-3mm}
%
%\caption{Visualization of Reconstructed Data (CESM:FREQSH)}
%\label{fig:vis-freqsh}
%\end{figure}

\section{Conclusion and Future Work}
\label{sec:conclusion}

In this paper, we explored leveraging convolutional autoencoders to improve error-bounded lossy compression. To this end, we developed an efficient method called AE-SZ, by integrating autoencoders in the SZ compression model with a series of optimizations. We comprehensively evaluated AE-SZ by comparing it with six related works on five real-world simulation datasets, with the following key findings. 
\begin{itemize}
    \item AE-SZ is  competitive in the low-bit-rate range (i.e., high-compression-ratio cases). Specifically, it exhibits the best rate-distortion results in 2D datasets. On 3D datasets, it obtains a much better rate-distortion than SZ2.1 and ZFP do (about 100\%$\sim$800\% improvement with the same PSNR). AE-SZ also exhibits very close rate distortions with those of SZinterp in high compression cases, demonstrating its great potential in error-bounded lossy compression.
    \item AE-SZ has a higher visual quality at the same compression ratio compared with SZauto, SZ2.1, and ZFP.
    \item AE-SZ is  slower than SZ2.1 and ZFP, 
    %GAIL - how much slower
    but is 30$\times$$\sim$200$\times$ faster than other autoencoder-based error-bounded lossy compressors.
\end{itemize} 

In the future, we plan to improve AE-SZ in several ways, including (1) optimizing the network structure and the hyperparameters of autoencoders in AE-SZ and 
%(2) exploring the possibilities of combining more types of traditional predictors with autoencoders.
%(3) designing more computational effective methods for selection between predictors; 
(2) speeding up the compression and decompression speeds for AE-SZ.

\section*{Acknowledgments}
This research was supported by the Exascale Computing Project (ECP), Project Number: 17-SC-20-SC, a collaborative effort of two DOE organizations – the Office of Science and the National Nuclear Security Administration, responsible for the planning and preparation of a capable exascale ecosystem, including software, applications, hardware, advanced system engineering and early testbed platforms, to support the nation’s exascale computing imperative. The material was supported by the U.S. Department of Energy, Office of Science, under contract DE-AC02-06CH11357, and supported by the National Science Foundation under Grant OAC-2003709 and OAC-2003624/2042084. We acknowledge the computing resources provided on Bebop (operated by Laboratory Computing Resource Center at Argonne) and on Theta and JLSE (operated by Argonne Leadership Computing Facility).

\bibliographystyle{IEEEtran}
\bibliography{references}

\begin{thebibliography}{10}
\providecommand{\url}[1]{#1}
\csname url@samestyle\endcsname
\providecommand{\newblock}{\relax}
\providecommand{\bibinfo}[2]{#2}
\providecommand{\BIBentrySTDinterwordspacing}{\spaceskip=0pt\relax}
\providecommand{\BIBentryALTinterwordstretchfactor}{4}
\providecommand{\BIBentryALTinterwordspacing}{\spaceskip=\fontdimen2\font plus
\BIBentryALTinterwordstretchfactor\fontdimen3\font minus
  \fontdimen4\font\relax}
\providecommand{\BIBforeignlanguage}[2]{{%
\expandafter\ifx\csname l@#1\endcsname\relax
\typeout{** WARNING: IEEEtran.bst: No hyphenation pattern has been}%
\typeout{** loaded for the language `#1'. Using the pattern for}%
\typeout{** the default language instead.}%
\else
\language=\csname l@#1\endcsname
\fi
#2}}
\providecommand{\BIBdecl}{\relax}
\BIBdecl

\bibitem{lcls-ii}
{SLAC National Accelerator Laboratory}, ``Linac coherent light source
  (lcls-ii),'' \url{https://lcls.slac.stanford.edu/}, 2017, online.

\bibitem{APSU}
T.~E. Fornek, ``{Advanced Photon Source Upgrade Project} preliminary design
  report,'' 9 2017.

\bibitem{use-case}
F.~Cappello, S.~Di, S.~Li, X.~Liang, G.~M. Ali, D.~Tao, C.~Yoon~Hong, X.-c. Wu,
  Y.~Alexeev, and T.~F. Chong, ``Use cases of lossy compression for
  floating-point data in scientific datasets,'' \emph{International Journal of
  High Performance Computing Applications (IJHPCA)}, vol.~33, pp. 1201--1220,
  2019.

\bibitem{gzip}
L.~P. Deutsch, ``{GZIP} file format specification version 4.3,'' 1996.

\bibitem{zstd}
Y.~Collet, ``Zstandard -- real-time data compression algorithm,''
  \emph{http://facebook.github.io/zstd/}, 2015.

\bibitem{zlib}
Zlib, \url{https://www.zlib.net/}, online.

\bibitem{fpc}
M.~{Burtscher} and P.~{Ratanaworabhan}, ``{FPC}: A high-speed compressor for
  double-precision floating-point data,'' \emph{IEEE Transactions on
  Computers}, vol.~58, no.~1, pp. 18--31, Jan 2009.

\bibitem{sz16}
S.~Di and F.~Cappello, ``Fast error-bounded lossy {HPC} data compression with
  {SZ},'' in \emph{IEEE International Parallel and Distributed Processing
  Symposium}, 2016, pp. 730--739.

\bibitem{sz17}
D.~Tao, S.~Di, Z.~Chen, and F.~Cappello, ``Significantly improving lossy
  compression for scientific data sets based on multidimensional prediction and
  error-controlled quantization,'' in \emph{2017 IEEE International Parallel
  and Distributed Processing Symposium}.\hskip 1em plus 0.5em minus 0.4em\relax
  IEEE, 2017, pp. 1129--1139.

\bibitem{cesm-eval-hpdf14}
A.~H. Baker, H.~Xu, J.~M. Dennis, M.~N. Levy, D.~Nychka, S.~A. Mickelson,
  J.~Edwards, M.~Vertenstein, and A.~Wegener, ``A methodology for evaluating
  the impact of data compression on climate simulation data,'' in
  \emph{Proceedings of the 23rd International Symposium on High-performance
  Parallel and Distributed Computing}, ser. HPDC '14.\hskip 1em plus 0.5em
  minus 0.4em\relax NY, USA: ACM, 2014, pp. 203--214.

\bibitem{ssem}
N.~Sasaki, K.~Sato, T.~Endo, and S.~Matsuoka, ``Exploration of lossy
  compression for application-level checkpoint/restart,'' in \emph{2015 IEEE
  International Parallel and Distributed Processing Symposium}.\hskip 1em plus
  0.5em minus 0.4em\relax IEEE, 2015, pp. 914--922.

\bibitem{Baker-Climate17}
A.~H. Baker, H.~Xu, D.~M. Hammerling, S.~Li, and J.~P. Clyne, ``Toward a
  multi-method approach: Lossy data compression for climate simulation data,''
  in \emph{High Performance Computing}.\hskip 1em plus 0.5em minus 0.4em\relax
  Springer International Publishing, 2017, pp. 30--42.

\bibitem{zfp}
P.~Lindstrom, ``Fixed-rate compressed floating-point arrays,'' \emph{IEEE
  transactions on visualization and computer graphics}, vol.~20, no.~12, pp.
  2674--2683, 2014.

\bibitem{Kai-HPDC2020}
K.~Zhao \emph{et~al.}, ``Significantly improving lossy compression for {HPC}
  datasets with second-order prediction and parameter optimization,'' in
  \emph{Proceedings of the 29th International Symposium on High-Performance
  Parallel and Distributed Computing}, ser. HPDC '20, 2020, pp. 89--100.

\bibitem{Xin-bigdata18}
X.~Liang, S.~Di, D.~Tao, S.~Li, S.~Li, H.~Guo, Z.~Chen, and F.~Cappello,
  ``Error-controlled lossy compression optimized for high compression ratios of
  scientific datasets,'' in \emph{2018 {IEEE} International Conference on Big
  Data}.\hskip 1em plus 0.5em minus 0.4em\relax IEEE, 2018.

\bibitem{son2014data}
S.~W. Son, Z.~Chen, W.~Hendrix, A.~Agrawal, W.-k. Liao, and A.~Choudhary,
  ``Data compression for the exascale computing era-survey,''
  \emph{Supercomputing frontiers and innovations}, vol.~1, no.~2, pp. 76--88,
  2014.

\bibitem{lindstrom2006fast}
P.~Lindstrom and M.~Isenburg, ``Fast and efficient compression of
  floating-point data,'' \emph{IEEE transactions on visualization and computer
  graphics}, vol.~12, no.~5, pp. 1245--1250, 2006.

\bibitem{wavesz}
J.~Tian, S.~Di, C.~Zhang, X.~Liang, S.~Jin, D.~Cheng, D.~Tao, and F.~Cappello,
  ``Wavesz: A hardware-algorithm co-design of efficient lossy compression for
  scientific data,'' in \emph{Proceedings of the 25th ACM SIGPLAN Symposium on
  Principles and Practice of Parallel Programming}, ser. PPoPP '20.\hskip 1em
  plus 0.5em minus 0.4em\relax New York, NY, USA: Association for Computing
  Machinery, 2020, p. 74–88.

\bibitem{ainsworth2019multilevel}
M.~Ainsworth, O.~Tugluk, B.~Whitney, and S.~Klasky, ``Multilevel techniques for
  compression and reduction of scientific data---the multivariate case,''
  \emph{SIAM Journal on Scientific Computing}, vol.~41, no.~2, pp.
  A1278--A1303, 2019.

\bibitem{li2017achieving}
S.~Li, N.~Marsaglia, V.~Chen, C.~M. Sewell, J.~P. Clyne, and H.~Childs,
  ``Achieving portable performance for wavelet compression using data parallel
  primitives,'' in \emph{EGPGV}, 2017, pp. 73--81.

\bibitem{delaunay2019evaluation}
X.~Delaunay, A.~Courtois, and F.~Gouillon, ``Evaluation of lossless and lossy
  algorithms for the compression of scientific datasets in {netCDF-4 or HDF5
  files},'' \emph{Geoscientific Model Development}, vol.~12, no.~9, pp.
  4099--4113, 2019.

\bibitem{zender2016bit}
C.~S. Zender, ``Bit grooming: statistically accurate precision-preserving
  quantization with compression, evaluated in the {netCDF Operators (NCO, v4.
  4.8+)},'' \emph{Geoscientific Model Development}, vol.~9, no.~9, pp.
  3199--3211, 2016.

\bibitem{lakshminarasimhan2013isabela}
S.~Lakshminarasimhan, N.~Shah, S.~Ethier, S.-H. Ku, C.-S. Chang, S.~Klasky,
  R.~Latham, R.~Ross, and N.~F. Samatova, ``{ISABELA} for effective in situ
  compression of scientific data,'' \emph{Concurrency and Computation: Practice
  and Experience}, vol.~25, no.~4, pp. 524--540, 2013.

\bibitem{quantum}
X.-C. Wu, S.~Di, E.~M. Dasgupta, F.~Cappello, H.~Finkel, Y.~Alexeev, and F.~T.
  Chong, ``Full-state quantum circuit simulation by using data compression,''
  in \emph{Proceedings of the International Conference for High Performance
  Computing, Networking, Storage and Analysis}, ser. SC '19.\hskip 1em plus
  0.5em minus 0.4em\relax New York, NY, USA: Association for Computing
  Machinery, 2019.

\bibitem{Pastri}
A.~M. {Gok}, S.~{Di}, Y.~{Alexeev}, D.~{Tao}, V.~{Mironov}, X.~{Liang}, and
  F.~{Cappello}, ``{PaSTRI}: Error-bounded lossy compression for two-electron
  integrals in quantum chemistry,'' in \emph{2018 IEEE International Conference
  on Cluster Computing (CLUSTER)}, Sep. 2018, pp. 1--11.

\bibitem{sz-psnr}
D.~{Tao}, S.~{Di}, X.~{Liang}, Z.~{Chen}, and F.~{Cappello}, ``Fixed-psnr lossy
  compression for scientific data,'' in \emph{2018 IEEE International
  Conference on Cluster Computing (CLUSTER)}, 2018, pp. 314--318.

\bibitem{cusz}
J.~Tian \emph{et~al.}, ``{CuSZ}: An efficient gpu-based error-bounded lossy
  compression framework for scientific data,'' in \emph{Proceedings of the ACM
  International Conference on Parallel Architectures and Compilation
  Techniques}, ser. PACT '20, 2020, p. 3–15.

\bibitem{8257962}
D.~Tao, S.~Di, Z.~Chen, and F.~Cappello, ``In-depth exploration of
  single-snapshot lossy compression techniques for n-body simulations,'' in
  \emph{2017 IEEE International Conference on Big Data (Big Data)}, 2017, pp.
  486--493.

\bibitem{DeepSZ}
S.~Jin, S.~Di, X.~Liang, J.~Tian, D.~Tao, and F.~Cappello, ``Deepsz: A novel
  framework to compress deep neural networks by using error-bounded lossy
  compression,'' in \emph{Proceedings of the 28th International Symposium on
  High-Performance Parallel and Distributed Computing}, ser. HPDC '19.\hskip
  1em plus 0.5em minus 0.4em\relax New York, NY, USA: ACM, 2019, pp. 159--170.

\bibitem{liang2018efficient}
X.~Liang, S.~Di, D.~Tao, Z.~Chen, and F.~Cappello, ``An efficient
  transformation scheme for lossy data compression with point-wise relative
  error bound,'' in \emph{2018 IEEE International Conference on Cluster
  Computing (CLUSTER)}.\hskip 1em plus 0.5em minus 0.4em\relax IEEE, 2018, pp.
  179--189.

\bibitem{sz_interp}
K.~Zhao, S.~Di, M.~Dmitriev, T.-L.~D. Tonellot, Z.~Chen, and F.~Cappello,
  ``Optimizing error-bounded lossy compression for scientiﬁc data by dynamic
  spline interpolation,'' in \emph{37th {IEEE} International Conference on Data
  Engineering}, 2021.

\bibitem{z-checker}
D.~Tao, S.~Di, H.~Guo, Z.~Chen, and F.~Cappello, ``Z-checker: A framework for
  assessing lossy compression of scientific data,'' \emph{The International
  Journal of High Performance Computing Applications}, vol.~33, no.~2, pp.
  285--303, 2019.

\bibitem{theis2017lossy}
L.~Theis, W.~Shi, A.~Cunningham, and F.~Husz{\'a}r, ``Lossy image compression
  with compressive autoencoders,'' \emph{arXiv preprint arXiv:1703.00395},
  2017.

\bibitem{cheng2018deep}
Z.~Cheng, H.~Sun, M.~Takeuchi, and J.~Katto, ``Deep convolutional
  autoencoder-based lossy image compression,'' in \emph{2018 Picture Coding
  Symposium (PCS)}.\hskip 1em plus 0.5em minus 0.4em\relax IEEE, 2018, pp.
  253--257.

\bibitem{zhou2018variational}
L.~Zhou, C.~Cai, Y.~Gao, S.~Su, and J.~Wu, ``Variational autoencoder for low
  bit-rate image compression,'' in \emph{Proceedings of the IEEE Conference on
  Computer Vision and Pattern Recognition Workshops}, 2018, pp. 2617--2620.

\bibitem{chen2019neural}
T.~Chen, H.~Liu, Z.~Ma, Q.~Shen, X.~Cao, and Y.~Wang, ``Neural image
  compression via non-local attention optimization and improved context
  modeling,'' \emph{arXiv preprint arXiv:1910.06244}, 2019.

\bibitem{balle2017endtoend}
J.~Ballé, V.~Laparra, and E.~P. Simoncelli, ``End-to-end optimized image
  compression,'' 2017.

\bibitem{balle2018variational}
J.~Balle, D.~Minnen, S.~Singh, S.~J. Hwang, and N.~Johnston, ``Variational
  image compression with a scale hyperprior,'' \emph{arXiv preprint
  arXiv:1802.01436}, 2018.

\bibitem{johnston2019computationally}
N.~Johnston, E.~Eban, A.~Gordon, and J.~Ball{\'e}, ``Computationally efficient
  neural image compression,'' \emph{arXiv preprint arXiv:1912.08771}, 2019.

\bibitem{glaws2020deep}
A.~Glaws, R.~King, and M.~Sprague, ``Deep learning for in situ data compression
  of large turbulent flow simulations,'' \emph{Physical Review Fluids}, vol.~5,
  no.~11, p. 114602, 2020.

\bibitem{he2016deep}
K.~He, X.~Zhang, S.~Ren, and J.~Sun, ``Deep residual learning for image
  recognition,'' in \emph{Proceedings of the IEEE conference on computer vision
  and pattern recognition}, 2016, pp. 770--778.

\bibitem{choineural}
J.~Choi, Q.~Gong, D.~Pugmire, S.~Klasky, M.~Churchill, S.-H. Ku, C.~Chang,
  J.~Lee, A.~Rangarajan, and S.~Ranka, ``Neural data compression for physics
  plasma simulation.''

\bibitem{liu2021high}
T.~Liu, J.~Wang, Q.~Liu, S.~Alibhai, T.~Lu, and X.~He, ``High-ratio lossy
  compression: Exploring the autoencoder to compress scientific data,''
  \emph{IEEE Transactions on Big Data}, 2021.

\bibitem{kolouri2018sliced}
S.~Kolouri, P.~E. Pope, C.~E. Martin, and G.~K. Rohde, ``Sliced {Wasserstein}
  auto-encoders,'' in \emph{International Conference on Learning
  Representations}, 2018.

\bibitem{tolstikhin2017wasserstein}
I.~Tolstikhin, O.~Bousquet, S.~Gelly, and B.~Schoelkopf, ``Wasserstein
  auto-encoders,'' \emph{arXiv preprint arXiv:1711.01558}, 2017.

\bibitem{lorenzo}
L.~Ibarria, P.~Lindstrom, J.~Rossignac, and A.~Szymczak, ``Out-of-core
  compression and decompression of large n-dimensional scalar fields,'' in
  \emph{Computer Graphics Forum}, vol.~22, no.~3.\hskip 1em plus 0.5em minus
  0.4em\relax Wiley Online Library, 2003, pp. 343--348.

\bibitem{balle2016density}
J.~Balle, V.~Laparra, and E.~P. Simoncelli, ``Density modeling of images using
  a generalized normalization transformation,'' 2016.

\bibitem{balle2018efficient}
J.~Balle, ``Efficient nonlinear transforms for lossy image compression,'' 2018.

\bibitem{maas2013rectifier}
A.~L. Maas, A.~Y. Hannun, and A.~Y. Ng, ``Rectifier nonlinearities improve
  neural network acoustic models,'' in \emph{Proc. icml}, vol.~30, no.~1.\hskip
  1em plus 0.5em minus 0.4em\relax Citeseer, 2013, p.~3.

\bibitem{ioffe2015batch}
S.~Ioffe and C.~Szegedy, ``Batch normalization: Accelerating deep network
  training by reducing internal covariate shift,'' in \emph{International
  conference on machine learning}.\hskip 1em plus 0.5em minus 0.4em\relax PMLR,
  2015, pp. 448--456.

\bibitem{jia2019layered}
C.~Jia, Z.~Liu, Y.~Wang, S.~Ma, and W.~Gao, ``Layered image compression using
  scalable auto-encoder,'' in \emph{2019 IEEE Conference on Multimedia
  Information Processing and Retrieval (MIPR)}.\hskip 1em plus 0.5em minus
  0.4em\relax IEEE, 2019, pp. 431--436.

\bibitem{xiao2019image}
L.~Xiao, H.~Wang, and N.~Ling, ``Image compression with deeper learned
  transformer,'' in \emph{2019 Asia-Pacific Signal and Information Processing
  Association Annual Summit and Conference (APSIPA ASC)}.\hskip 1em plus 0.5em
  minus 0.4em\relax IEEE, 2019, pp. 53--57.

\bibitem{kingma2013auto}
D.~P. Kingma and M.~Welling, ``Auto-encoding variational bayes,'' \emph{arXiv
  preprint arXiv:1312.6114}, 2013.

\bibitem{higgins2016beta}
I.~Higgins, L.~Matthey, A.~Pal, C.~Burgess, X.~Glorot, M.~Botvinick,
  S.~Mohamed, and A.~Lerchner, ``beta-{VAE}: Learning basic visual concepts
  with a constrained variational framework,'' 2016.

\bibitem{kumar2018dipvae}
A.~Kumar, P.~Sattigeri, and A.~Balakrishnan, ``Variational inference of
  disentangled latent concepts from unlabeled observations,'' 2018.

\bibitem{zhao2018infovae}
S.~Zhao, J.~Song, and S.~Ermon, ``{InfoVAE}: Information maximizing variational
  autoencoders,'' 2018.

\bibitem{chen2018logcosh}
P.~Chen, G.~Chen, and S.~Zhang, ``Log hyperbolic cosine loss improves
  variational auto-encoder,'' 2018.

\bibitem{sdrb}
K.~Zhao, S.~Di, X.~Lian, S.~Li, D.~Tao, J.~Bessac, Z.~Chen, and F.~Cappello,
  ``{SDRBench}: Scientific data reduction benchmark for lossy compressors,'' in
  \emph{2020 IEEE International Conference on Big Data (Big Data)}, 2020, pp.
  2716--2724.

\bibitem{cesm}
J.~E. Kay and et\ al., ``The {Community Earth System Model (CESM)} large
  ensemble project: A community resource for studying climate change in the
  presence of internal climate variability,'' \emph{Bulletin of the American
  Meteorological Society}, vol.~96, no.~8, pp. 1333--1349, 2015.

\bibitem{geodriveFirstBreak2020}
S.~Kayum \emph{et~al.}, ``{GeoDRIVE} -- a high performance computing flexible
  platform for seismic applications,'' \emph{First Break}, vol.~38, no.~2, pp.
  97--100, 2020.

\bibitem{nyx}
{NYX simulation}, \url{https://amrex-astro.github.io/Nyx}, 2019, online.

\bibitem{hurricane}
{Hurricane ISABEL simulation data},
  \url{http://vis.computer.org/vis2004contest/data.html}, 2004, online.

\bibitem{exafel}
E.~project,
  \url{https://www.exascaleproject.org/project/exafel-data-analytics-exascale-free-electron-lasers/},
  2019, online.

\bibitem{cappello2019use}
F.~Cappello, S.~Di, S.~Li, X.~Liang, A.~M. Gok, D.~Tao, C.~H. Yoon, X.-C. Wu,
  Y.~Alexeev, and F.~T. Chong, ``Use cases of lossy compression for
  floating-point data in scientific data sets,'' \emph{The International
  Journal of High Performance Computing Applications}, vol.~33, no.~6, pp.
  1201--1220, 2019.

\bibitem{liang2019significantly}
X.~Liang, S.~Di, S.~Li, D.~Tao, B.~Nicolae, Z.~Chen, and F.~Cappello,
  ``Significantly improving lossy compression quality based on an optimized
  hybrid prediction model,'' in \emph{Proceedings of the International
  Conference for High Performance Computing, Networking, Storage and Analysis},
  2019, pp. 1--26.

\end{thebibliography}

\end{document}